\documentclass[10pt,twocolumn,letterpaper]{article}

\usepackage[pagenumbers]{iccv} %

\definecolor{iccvblue}{rgb}{0.21,0.49,0.74}
\usepackage[pagebackref,breaklinks,colorlinks,allcolors=iccvblue]{hyperref}

\usepackage{graphicx}
\usepackage{booktabs}
\usepackage{wrapfig}
\usepackage{array}
\usepackage{tikz}
\usepackage{pgfplots}
\usepackage{colortbl}
\usepackage[accsupp]{axessibility}  %
\usepackage{orcidlink}
\usepackage{tabularx}
\usepackage{multirow}
\usepackage{tcolorbox}
\usepackage{tikz}
\usepackage{pifont}
\usepackage{microtype}
\usepackage{fontawesome}
\usepackage[switch]{lineno} %
\usepackage[toc,page]{appendix}
\usepackage{bm}
\usepackage{gradient-text}
\usepackage{textcomp}
\usepackage{gensymb}
\usepackage{lipsum}

\usepackage{caption}
\newcommand{\qheading}[1]{\noindent\mbox{\textbf{#1}}}
\newcommand{\pheading}[1]{\medskip\noindent\textbf{#1}}

\definecolor{bestgreen}{RGB}{153,200,76}
\definecolor{worstred}{RGB}{192,0,0}

\definecolor{cbad}{HTML}{FFD0D0} 
\definecolor{cmedium}{HTML}{FFF0D0}
\definecolor{cgood}{HTML}{90C060}

\usepackage{pgfplots}
\pgfplotsset{compat=1.18}

\newlength\savewidth\newcommand\shline{\noalign{\global\savewidth\arrayrulewidth
  \global\arrayrulewidth 1pt}\hline\noalign{\global\arrayrulewidth\savewidth}}

\definecolor{DeltaColor}{rgb}{0.039,0.73,0.71}
\definecolor{SigmaColor}{rgb}{0.98,0.45,0.0}
\definecolor{AlphaColor}{rgb}{0,0,0.8}
\definecolor{BetaColor}{rgb}{0.8,0,0.8}
\definecolor{GammaColor}{rgb}{0.514,0.34,0.224}
\definecolor{EpsilonColor}{rgb}{0.353,0.725,0.906}
\definecolor{PurpleColor}{HTML}{9839ff}
\definecolor{RedColor}{rgb}{0.949,0.275, 0.224}
\definecolor{citecolor}{HTML}{0071bc}

\newcommand{\page}{\href{https://boqian-li.github.io/ETCH/}{\textcolor{magenta}{\xspace\tt\textit{boqian-li.github.io/ETCH}}}\xspace}

\newcommand{\web}{\href{https://boqian-li.github.io/ETCH/}{\textcolor{magenta}{\xspace\tt{website}}}\xspace}

\newcommand{\modelname}{\mbox{\textcolor{black}{ETCH}}\xspace}

\newcommand{\longtitle}{Generalizing Body Fitting to Clothed Humans via Equivariant Tightness}
\newcommand{\ourtitle}{\modelname: \longtitle}

\newcommand{\xmark}{\textcolor{RedColor}{\ding{55}}\xspace}
\newcommand{\cmark}{\textcolor{GreenColor}{\ding{51}}\xspace}

\newcommand{\gt}{{ground-truth}\xspace}

\newcommand{\ood}{{out-of-distribution}\xspace}

\newcommand{\arteq}{\mbox{ArtEq}\xspace}
\newcommand{\ipnet}{\mbox{IPNet}\xspace}
\newcommand{\tightcap}{\mbox{TightCap}\xspace}

\newcommand{\ddress}{\mbox{4D-Dress}\xspace}

\newcommand{\smplx}{\mbox{SMPL-X}\xspace}
\newcommand{\smpl}{\mbox{SMPL}\xspace}

\newcommand{\sota}{state-of-the-art\xspace}

\newcommand{\real}{\mathbb{R}}
\newcommand{\vect}[1]{\mathbf{#1}}

\newcommand{\shapecoeff}{\boldsymbol{\beta}}
\newcommand{\shapedim}{{\left| \shapecoeff \right|}}

\newcommand{\posecoeff}{\boldsymbol{\theta}}

\newcommand{\template}{\mathbf{\bar{T}}}
\newcommand{\restpose}{\mathbf{T}}

\newcommand{\blendweights}{\mathcal{W}}

\definecolor{PurpleColor}{HTML}{8B008B}
\definecolor{OrangeColor}{rgb}{0.914,0.541,0.0.141}
\definecolor{GreenColor}{rgb}{0.137,0.573,0.565}

\newcommand*\fullcirc[1][0.5ex]{\tikz\fill (0,0) circle (#1);}

\title{\ourtitle}

\author{
Boqian Li$^{1}$ \quad Haiwen Feng$^{2,3*}$ \quad Zeyu Cai$^{1}$ \quad Michael J. Black$^{2}$ \quad Yuliang Xiu$^{1,2\dagger}$ \vspace{5pt}\\
$^1$Westlake University \quad $^2$Max Planck Institute for Intelligent Systems \quad
$^3$UC Berkeley \\
{
    \tt
    \small
    \{liboqian, caizeyu, xiuyuliang\}@westlake.edu.cn \quad \{haiwen.feng, black\}@tuebingen.mpg.de} \vspace{5pt}\\
\href{https://boqian-li.github.io/ETCH/}{\textcolor{magenta}{\texttt{\small boqian-li.github.io/ETCH}}}\\
}

\begin{document}

\newcommand{\teaserCaption}{\textbf{Body Fitting on Clothed Humans.} Given 3D clothed humans in any pose and clothing, \modelname accurately fits the \textit{body underneath}. Our key novelty is modeling cloth-to-body SE(3)-equivariant \gradientRGB{tightness vectors}{254,217,118}{192,50,26} for clothed humans, abbreviated as \modelname, which resembles ``etching'' from the outer clothing down to the inner body. The \gt body is shown in \textcolor{gtbody}{blue}, our fitted body in \textcolor{predbody}{green}, and \gt markers as \textcolor{red}{\fullcirc}. }

\definecolor{gtbody}{RGB}{89,147,203}
\definecolor{predbody}{RGB}{101,167,158}

\twocolumn[{
    \renewcommand\twocolumn[1][]{#1}
    \maketitle
    \centering
    \begin{minipage}{\textwidth}
        \centering
        \includegraphics[trim=000mm 000mm 000mm 000mm, clip=true, width=\linewidth]{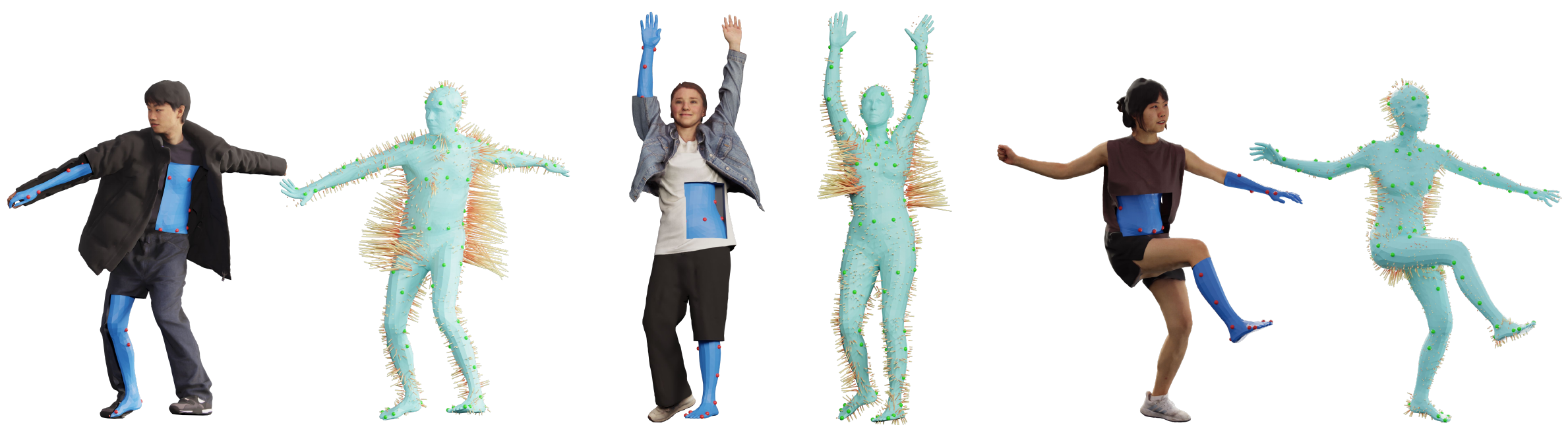}
    \end{minipage}
    \captionsetup{type=figure}
    \captionof{figure}{\teaserCaption}
    \label{fig:teaser}
    \vspace{1.2em}
}]

\def\thefootnote{*}\footnotetext{Project Lead}
\def\thefootnote{$\dagger$}\footnotetext{Corresponding Author}

\begin{abstract}

Fitting a body to a 3D clothed human point cloud is a common yet challenging task. Traditional optimization-based approaches use multi-stage pipelines that are sensitive to pose initialization, while recent learning-based methods often struggle with generalization across diverse poses and garment types.
We propose Equivariant Tightness Fitting for Clothed Humans, or \modelname, a novel pipeline that estimates cloth-to-body surface mapping through locally approximate SE(3) equivariance, encoding tightness as displacement vectors from the cloth surface to the underlying body. Following this mapping, pose-invariant body features regress sparse body markers, simplifying clothed human fitting into an inner-body marker fitting task.
Extensive experiments on CAPE and \ddress show that \modelname significantly outperforms \sota methods -- both tightness-agnostic and tightness-aware -- in body fitting accuracy on loose clothing (\textcolor{ForestGreen}{$16.7\% \sim 69.5\%$}) and shape accuracy (average \textcolor{ForestGreen}{$49.9\%$}). Our equivariant tightness design can even reduce directional errors by (\textcolor{ForestGreen}{$67.2\% \sim 89.8\%$}) in one-shot (or \ood) settings ($\approx1\%$ data). Qualitative results demonstrate strong generalization of \modelname, regardless of challenging poses, unseen shapes, loose clothing, and non-rigid dynamics. We will release the code and models soon for research purposes at \page.

\end{abstract}
    
\section{Introduction}
\label{sec:intro}

Fitting a template body mesh to a 3D clothed human point cloud is a long-standing challenge, which is vital for shape matching, motion capture, and animation, while enabling numerous applications like virtual try-on and immersive teleportation. It also supports the development of parametric human body models such as \smpl~\cite{SMPL:2015} and GHUM~\cite{xu2020ghum}. 
Aligning a template body mesh with raw human scans, \ie, unordered point clouds, provides a common topology supporting statistical modeling of body shape and deformation.

Traditional optimization-based fitting pipelines (details in \cref{subsec:reg_fit}) are complex, depending on many ad-hoc rules to address corner cases, such as inconsistent 2D keypoints and additional skin-clothing separation.
A path forward would leverage 3D scans paired with accurate underlying bodies to train a data-driven 3D pose regressor. 
However, despite advances in point-based neural networks, like PointNet~\cite{qi2017pointnet} and PointNet++~\cite{qi2017pointnet++}, they still struggle with fitting tasks, particularly in achieving \textbf{robust generalization} under the complexity of \textit{varied human poses and shapes, various clothing styles, and non-rigid dynamics}. 

In recent works on fitting a parametric body model to scans,
\arteq~\cite{feng2023generalizing} addresses \ood pose generalization through articulated SE(3) equivariance. However, while \arteq is designed to tackle articulation, it struggles with significant clothing deviations from the underlying body, particularly with loose garments like skirts or large deformations caused by dynamic motions, as shown in \cref{fig:teaser}.  %
Other recent approaches attempt to address this by separating garment layers from the underlying body, using scalar tightness~\cite{chen2021tightcap} or double-layer occupancy~\cite{bhatnagar2020ipnet,wang2021ptf}. These methods simplify fitting loosely clothed humans by reducing the problem to a bare-body fitting problem. 
However, given the combinatorial deformations caused by the variety of human motions and clothing, these methods still struggle with \ood poses or shapes, as illustrated in \cref{fig:comparison}.

We integrate the two -- ``equivariance'' and ``tightness'', capitalizing on their strengths to achieve the best of both worlds, and present \emph{Equivariant Tightness Fitting for Clothed Humans (\modelname)}. 
Specifically, \modelname encodes tightness as displacement vectors from the observed cloth surface to the underlying body surface, capturing their dynamic interaction. This encoding remains approximately locally SE(3) equivariant under clothing deformation, accommodates both tight- and loose-fitting garments, and consistently points toward the inner body regardless of pose and clothing deformations. The encoding is learned from  \gt tightness vectors computed from paired data, \ie, the clothed human meshes paired with the underlying \gt body.

\begin{figure}
    \captionsetup{labelfont={scriptsize}}
    \scriptsize
    \centering{
    \includegraphics[trim=000mm 000mm 000mm 000mm, clip=true, width=\linewidth]{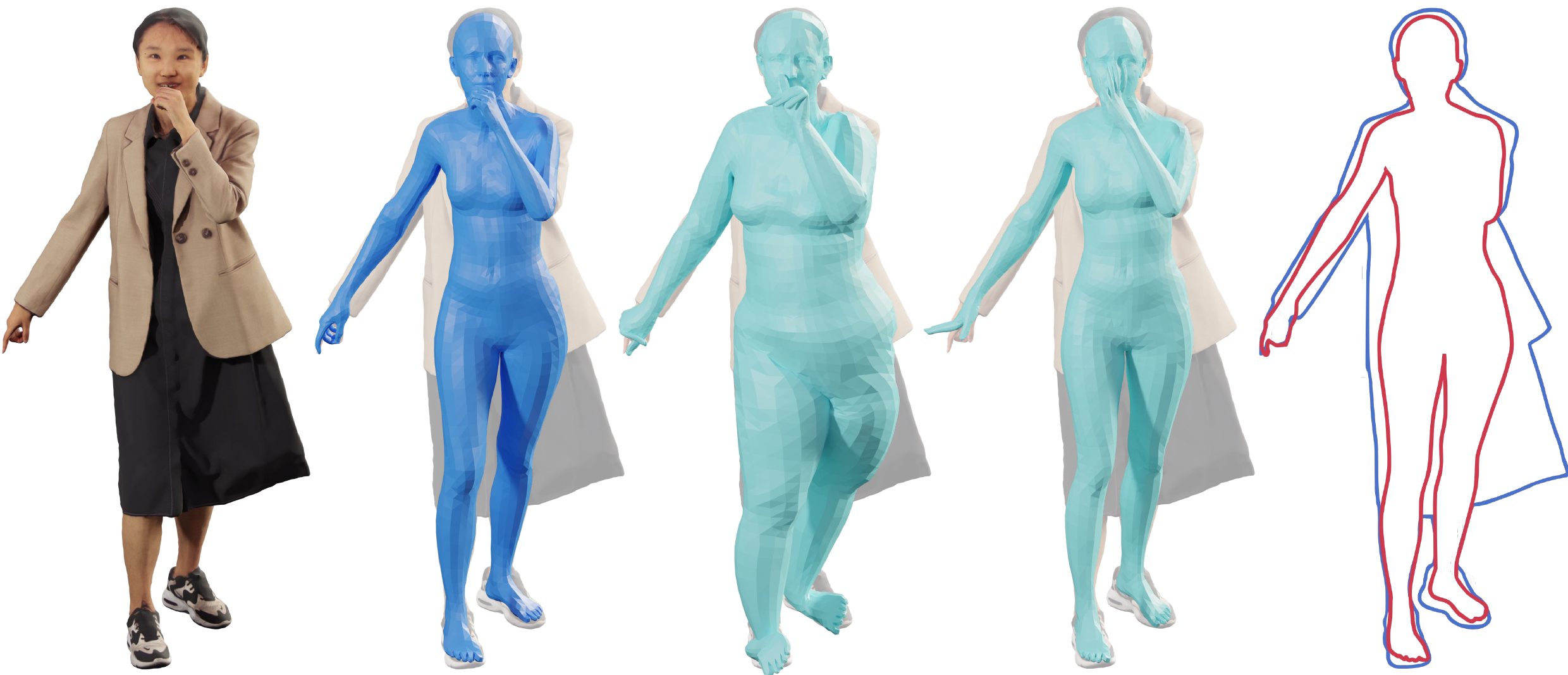}
    \begin{tabularx}{\linewidth}{
        >{\centering\arraybackslash}X
        >{\centering\arraybackslash}X
        >{\centering\arraybackslash}X
        >{\centering\arraybackslash}X
        >{\centering\arraybackslash}X} 
        Scan & GT body & NICP~\cite{marin24nicp} & Ours & \textcolor{red}{body}-\textcolor{blue}{cloth}
    \end{tabularx}
    \caption{\textbf{Registration vs. Fitting.} Though both registration and fitting involve placing body inside clothing, ``registration'', like NICP~\cite{marin24nicp}, focuses on matching the \textcolor{blue}{\textit{outer surface}}, whereas ``fitting'' emphasizes aligning with the \textcolor{red}{\textit{underlying body}}, making it more robust to clothing variations. 
    }
    \label{fig:reg_fit}}
\end{figure}

As shown in~\cref{fig:reg_fit}, during inference the predicted tightness vectors ($\mathbf{V}$) use the \textcolor{blue}{outer surface points} to approximate the \textcolor{red}{inner points} that define the body's shape. However, this only accomplishes half the task of fitting the body, as the points remain unordered and do not offer a direct correspondence. 
To ensure robust body fitting, we perform sparse marker regression, which is more efficient and less sensitive to outliers compared to dense correspondence methods like PTF~\cite{wang2021ptf} (see~\cref{tab:ablation}). In this approach, we aggregate the dense inner points into a sparse set of virtual markers on the body surface. As shown in~\cref{fig:tightness}, each marker is assigned a region label ($\mathbf{L}$) corresponding to a specific area of the body, and the dense inner points sharing the same label—those clustered within the same region—are weighted and aggregated based on their predicted confidence ($\mathbf{C}$). This process yields a robust localization of the markers on the body surface.
Consequently, fitting a parametric body to these markers is a well-solved problem \cite{Loper:SIGASIA:2014}. 

And unlike direct body regression methods like \arteq~\cite{feng2023generalizing}, our approach is tightness-aware. By correctly disentangling cloth deformation from body structure in an equivariant manner and regressing the sparse markers with invariant features, \modelname demonstrates strong generalization across challenging poses, extreme shapes, diverse garments, and clothing dynamics, supported by extensive qualitative and quantitative results. Each design choice has been validated through ablations.

Our key contributions are as follows:
\begin{itemize}
    \item \textbf{Equivariant Tightness Fitting.} We introduce a novel framework, abbreviated as \modelname, which models cloth-to-body mapping using \emph{tightness-equivariant displacement vectors} and leverages \emph{pose-invariant body correspondences} and \emph{weighted aggregation} for accurate sparse marker placement. This reformulates the clothed human fitting into a tightness-aware marker-based fitting task.
    \item \textbf{Superior Performance.} \modelname significantly outperforms prior approaches (\ie, \ipnet, PTF, NICP, \arteq) across datasets and metrics (see \cref{tab:benchmark}). It shows superior \emph{generalization} to loose dynamic clothing and challenging poses and extreme shapes at \cref{fig:comparison,fig:consistency}, and shows strong \ood generalization at \cref{fig:equiv,tab:equiv}.
    \item \textbf{In-depth Analysis.} We perform extensive ablation studies to assess each component's contribution, including tightness (direction vs. scalar) at \cref{tab:ablation}, sparse markers vs. dense correspondences at \cref{tab:ablation,fig:marker_corr}, the role of equivariance features at \cref{fig:equiv,tab:equiv}, and the effectiveness of post-refinement at ~\cref{tab:chamfer}.
\end{itemize}

\section{Related Work}
\label{sec:related}

\subsection{Body Fitting for 3D Clothed Humans}
\label{subsec:reg_fit}

Estimating body shape under clothing has a long history.
Balan and Black \cite{Balan:ECCV} introduce the problem of fitting a parametric 3D human body model {\em inside} a clothed 3D surface.
They work with a coarse visual-hull representation extracted from multi-view images or video while later work uses 3D scans \cite{WUHRER201431,HASLER2009211,wide2016,Song2016}.
All these methods rely on the same principle of optimizing for a single parametric body shape that, when posed, fits inside all the scans.
None of them explicitly model how clothing deviates from the body.
The first work to learn a mapping between the inner body shape and outer, clothed, shape does so in 2D \cite{guan20102d} by building a statistical model of how clothing deviates from the body.

\pheading{Surface Registration vs. Body Fitting.} 
Modeling the cloth-to-body mapping relates to scan fitting terminology. ``Surface Registration'' and ``Body Fitting'' are often used interchangeably but describe distinct processes. As illustrated in~\cref{fig:reg_fit}, for 3D clothed humans, body fitting aims to deform a body model (\eg, SMPL~\cite{SMPL:2015}) to align with the \textcolor{red}{\textit{underlying body}}, optimizing within the model space (\eg, SMPL's pose and shape parameters). In contrast, registration (or ``alignment'') deforms a template mesh to match the \textcolor{blue}{\textit{outer surface}}, optimizing outside the model space (\eg, via displacements from the body, like SMPL-D~\cite{zhang2017buff,bhatnagar2020ipnet,ma2020cape,gerard2017clothcap}). These terms can be interchangeable when clothing is minimal~\cite{hirshberg2012coregistration,bogo2014faust,li08global}. However, for loose clothing, ``tightness-agnostic'' registration will inflate the human shapes to align with the outer clothing surface, as shown in~\cref{fig:reg_fit,fig:comparison}.

With \modelname, our goal is to achieve generalizable {\em body fitting} for 3D clothed humans by separating clothing from the underlying SMPL body shape. Our approach follows the ``tightness-aware'' paradigm (see~\cref{tab:benchmark}) established by \tightcap~\cite{chen2021tightcap}, \ipnet~\cite{bhatnagar2020ipnet}, and PTF~\cite{wang2021ptf}. 
Prior work has exploited the fact that the body should fit tightly to the scan in skin regions but more loosely in clothed regions \cite{zhang2017buff,patel2021agora}.
Zhang et al.~\cite{zhang2017buff} use a binary tightness model in which the body ``snaps'' to nearby clothing surfaces but ignores distant ones.
Neophytou and Hilton \cite{hilton2014} introduce a learned 3D model of clothing displacement from a parametric body shape, but this model has to be trained for each actor.
Our key innovation is defining tightness as a set of vectors, rather than relying on a scalar UV map~\cite{chen2021tightcap} or double-layer occupancy~\cite{bhatnagar2020ipnet}. Additionally, we conduct the fitting in the posed rather than canonical space like PTF~\cite{wang2021ptf}, being more robust to raw 3D human captures with \ood poses.

\pheading{Optimization vs. Learning.}
Prior work could also be grouped into ``Optimization-based'' and ``Learning-based'' approaches.
Optimization approaches refine results iteratively using the ICP algorithm and its variants~\cite{chen1992object,allen2003space,pons2015dyna,zuffi2015stitched}, often aided by additional cues such as markers~\cite{allen2003space,anguelov2005scape,pishchulin2017building} or color patterns~\cite{bogo2014faust,bogo2017dynamic}. However, their sensitivity to noise and poor initial poses can hinder convergence, resulting in suboptimal local minima or failed fitting.

In particular, the most commonly used optimization-based body fitting pipelines~\cite{zhang2017buff,ma2020cape,patel2021agora,zheng2019deephuman,tao2021function4d,rbh_reg} typically proceed as follows: multiview rendering of scans $\Rightarrow$ 2D keypoint estimation (\eg, OpenPose~\cite{cao2019openpose}, AlphaPose~\cite{fang2022alphapose}, \etc) $\Rightarrow$ triangulation of 3D keypoints from multiview 2D keypoints $\Rightarrow$ fitting SMPL (-X) parameters using 3D keypoints and the 3D point cloud (often accompanied by skin segmentation~\cite{antic2024close,Gong2019Graphonomy} or self-intersection penalties~\cite{mueller2021tuch}). 2D keypoint estimation may fail in loose clothing, and triangulation based on such inconsistent 2D keypoints can lead to skewed 3D joint estimates. This harms the fitting process, as chamfer-based optimization is highly sensitive to pose initialization and can easily get trapped in local minima.
 
Learning-based methods, powered by the large-scale 3D datasets (\eg, AMASS~\cite{mahmood2019amass}) and point cloud networks (\eg, PointNet~\cite{qi2017pointnet,qi2017pointnet++}, DeepSets~\cite{zaheer2017deep}, KPConv~\cite{thomas2019kpconv}, and PointTransformer~\cite{zhao2021point,wu2022point,wu2024point}), either improve initialization for subsequent optimization steps~\cite{wang2021ptf,bhatnagar2020ipnet}, directly output meshes~\cite{prokudin2019efficient,wang2020sequential,zhou2020reconstructing}, or directly predict the pose/shape parameters of statistical body models~\cite{bhatnagar2020loopreg,jiang2019skeleton,liu2021votehmr,zhou2020reconstructing}. 
However, direct model-based regression is difficult~\cite{bhatnagar2020ipnet,bhatnagar2020loopreg,wang2021ptf,jiang2019skeleton,wang2020sequential}. To circumvent this, current approaches use intermediate representations, such as joint features~\cite{jiang2019skeleton,liu2021votehmr}, correspondence maps~\cite{bhatnagar2020loopreg,wang2021ptf}, segmentation parts~\cite{wang2021ptf,feng2023generalizing,bhatnagar2020ipnet,bhatnagar2020loopreg,feng2023generalizing}, or resort to temporal consistency~\cite{zhang2017buff} or motion models~\cite{jiang2022h4d}.

\subsection{SE(3)Equivariant and Human Pointcloud}
Convolutional neural networks (CNNs) owe much of their success to translational equivariance. Building on this, researchers have explored SE(3)-equivariant neural networks for point cloud processing, including Vector Neurons~\cite{deng2021vector}, Tensor Field Networks~\cite{thomas2018tensor}, SE3-transformers~\cite{fuchs2020se}, and methods leveraging group averaging theory or SO(3) discretization. For example, EPN~\cite{chen2021equivariant} discretizes SO(3) and applies separable discrete convolutions (SPConv) for computational efficiency, simplifying rigid-body pose regression.

Extending to non-rigid scenarios, recent work addresses part-level articulation at the object or scene level (\eg, \cite{Deng2023BananaBF, Zhong2023MultibodySE,Puny2021FrameAF}). Human bodies represent a special case due to their high articulation. Efforts like Articulated SE(3) Equivariance (\arteq)~\cite{feng2023generalizing} or  APEN~\cite{atzmon2024approximately} model local or piecewise SO(3) equivariance to handle articulated deformations. However, while the human skeleton is articulated, loose clothing does not strictly follow these assumptions: garments can undergo large, complex deformations that violate standard rigidity or articulation models. Consequently, achieving robust body fitting via approximate SE(3) equivariance for loosely clothed humans remains an open challenge.

In contrast, we address this challenge from a new perspective by modeling approximate equivariance between the cloth surface and body surface through the local SO(3) equivariance of a ``tightness'' vector. This approach enables explicit cloth-to-body modeling while taking advantage of approximate equivariance to achieve more robust body fitting under loose clothing, effectively combining the strengths of learning-based pointcloud-to-marker regression and optimization-based marker-to-body fitting.

\section{Method}
\label{sec:method}

\captionsetup{labelfont={scriptsize}}

\begin{figure*}[t]
\scriptsize
    \centering{
    \includegraphics[trim=000mm 000mm 000mm 000mm, clip=true, width=0.8\linewidth]{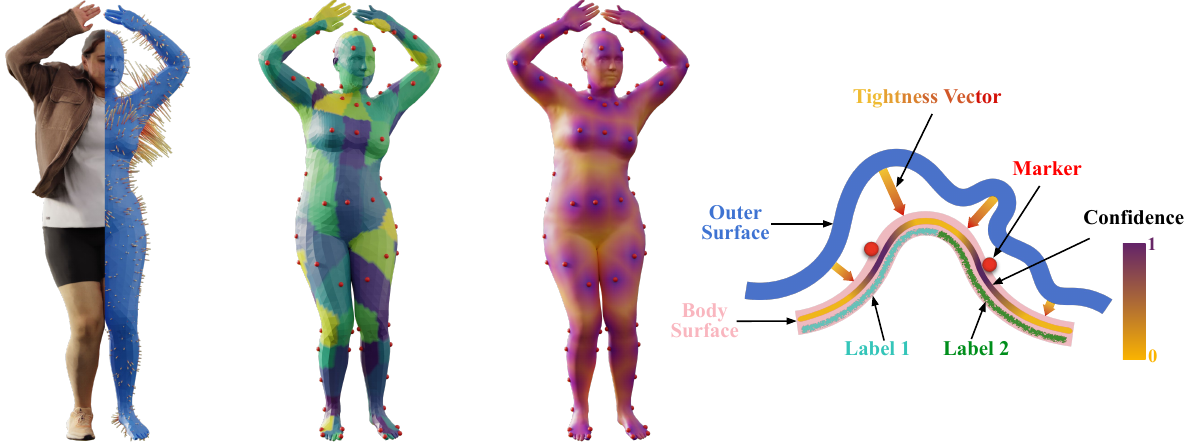}
    \begin{tabularx}{0.8\linewidth}{
         >{\centering\arraybackslash}X
         >{\centering\arraybackslash}X
         >{\centering\arraybackslash}X
         >{\centering\arraybackslash}p{5.0cm}
        } Tightness Vector & Marker-based Labels & Geodesic-based Confidence & Unified 2D Illustration \\
        $\mathbf{V}$ & $\mathbf{L}$ & $\mathbf{C}$ & 
     \end{tabularx}
    \caption{\textbf{Terminology of Tightness-Vector and Marker-Confidence.} We illustrate the key components used for data preparation: 1) Tightness Vectors $\mathbf{V}$, which connect the outer surface points with underneath body, and transmitting 2) Marker-based Labels $\mathbf{L}$ and Confidence $\mathbf{C}$. We also provide a 2D illustration that unifies these terms together. Sparse markers as \textcolor{red}{\fullcirc}, and \gradientRGB{confidence bar}{242,171,8}{105,41,100} indicates the geodesic distance to the closest marker.}
    \label{fig:tightness}
    }
\end{figure*}

Given a point cloud $\vect{X} = \{ \vect{x}_i \in \real^3 \}_N$, which is randomly sampled from the 3D humans, our goal is to optimize the pose $\posecoeff$, shape $\shapecoeff$ and $\mathbf{t}$ parameters of SMPL~\cite{SMPL:2015} (\cref{subsec:preliminary}), via a proxy task -- fitting SMPL to the estimated sparse body markers, denoted as $\hat{\vect{M}} = \{\hat{\vect{m}_k} \in \real^3 \}_K, \text{here } K=86$. 
To accurately estimate these body markers, we leverage the SE(3)-equivariant \& invariant pointwise features (\cref{subsec:preliminary}). 
Using these pointwise features as input, we regress three key components to model the \emph{cloth-to-body tightness}, refer to~\cref{fig:tightness} for illustrations: \textbf{1) a vector $\mathbf{v}_i$} -- pointing from the outer surface point cloud to the inner body surface (\cref{subsec:vector}), constructed with the directional term $\mathbf{d}_i$ and magnitude $b_i=\| \mathbf{v}_i \|$, \textbf{2) a label $l_i$} -- indicating the corresponding inner marker of $\mathbf{x}_i$, \textbf{3) a confidence $c_i$} -- quantifies the contribution of inner points $\mathbf{y}_i$ in marker aggregation (\cref{subsec:label-confidence}). 
With these elements, we apply weighted (with estimated confidence) aggregation with tightness vector clusters of each marker label, to accurately regress the ultimate body markers $\hat{\mathbf{M}}$, which are finally used to optimize the final body model (\cref{subsec:marker-smpl}).
How to prepare the \gt components $\{\mathbf{V,L,C}\}$ for training is detailed in~\cref{subsec:data}.

\subsection{Preliminary}
\label{subsec:preliminary}

\qheading{Parametric Body Model -- SMPL.}
SMPL~\cite{SMPL:2015} has been a standard body format in various clothed human datasets~\cite{ma2020cape,wang20244ddress,zheng2019deephuman,tao2021function4d}. It is a statistical model that maps shape $\shapecoeff \in \real^{10}$ and pose $\posecoeff \in \real^{J \times 3}$ parameters to mesh vertices $\vect{V} \in \real^{6890 \times 3}$, where $J$ is the number of human joints ($J=24$). 
$\shapecoeff$ contains the blend weights of shape blendshapes $B_S$, and $B_S(\shapecoeff)$ accounts for variations of body shapes. $\posecoeff$ contains the relative rotation (axis-angle) of each joint plus the root one \wrt their parent in the kinematic tree, and $B_P(\posecoeff)$ models the pose-dependent deformation. Both shape displacements $B_S(\shapecoeff)$ and pose correctives $B_P(\posecoeff)$ are added together onto the template mesh $\template \in \real^{6890 \times 3}$, in the rest pose (or T-pose), to produce the output mesh $\restpose$: 

\begin{equation}
    \restpose(\shapecoeff, \posecoeff) = \template + B_S(\shapecoeff) + B_P(\posecoeff), 
\end{equation}
Next, the joint regressor $J(\shapecoeff)$ is applied to the rest-pose mesh $\restpose$ to obtain the 3D joints : $\real^{\shapedim} \to \real^{J \times 3}$. Finally, Linear Blend Skinning (LBS) $W(\cdot)$ is used for reposing purposes, the skinning weights are denoted as $\blendweights$, then the posed mesh is translated with $\boldsymbol{t} \in \real^3$ as final output $\mathbf{M}$ :

\begin{equation}
\label{eq:smpl}
    \mathbf{M}(\shapecoeff, \posecoeff, \mathbf{t}) = W(
    \restpose(\shapecoeff, \posecoeff), J(\shapecoeff), \posecoeff, \boldsymbol{t}, \blendweights ). %
\end{equation}

\qheading{Local SE(3) Equivariance Features.}
\label{ssec:preliminaries_so3}
A function $\mathcal{F}$ is equivariant if, for any transformation $\mathcal{T}$, $\mathcal{F}(\mathcal{T}\mathbf{X})=\mathcal{T}\mathcal{F}(\mathbf{X})$, this property aids in training within only ``canonical'' space $\overline{\mathcal{T}}\mathcal{F}(\mathbf{X})$, while generalizing to any $\mathcal{T}\mathcal{F}(\mathbf{X})$.
Gauge-equivariant neural networks extend 2D convolutions to manifolds by ``shifting'' kernels across tangent frames to achieve gauge symmetry equivariance~\cite{cohen2019gauge}. Due to the high computational cost, the continuous $\mathrm{SO}(3)$ space is approximated using the icosahedron’s 60 rotational symmetries. EPN~\cite{chen2021equivariant} further extended this idea to point clouds and $\mathrm{SE}(3)$ by introducing separable convolutional layers that handle rotations and translations independently.

\begin{figure}
\captionsetup{labelfont={scriptsize}}
\centering
\includegraphics[width=0.9\linewidth]{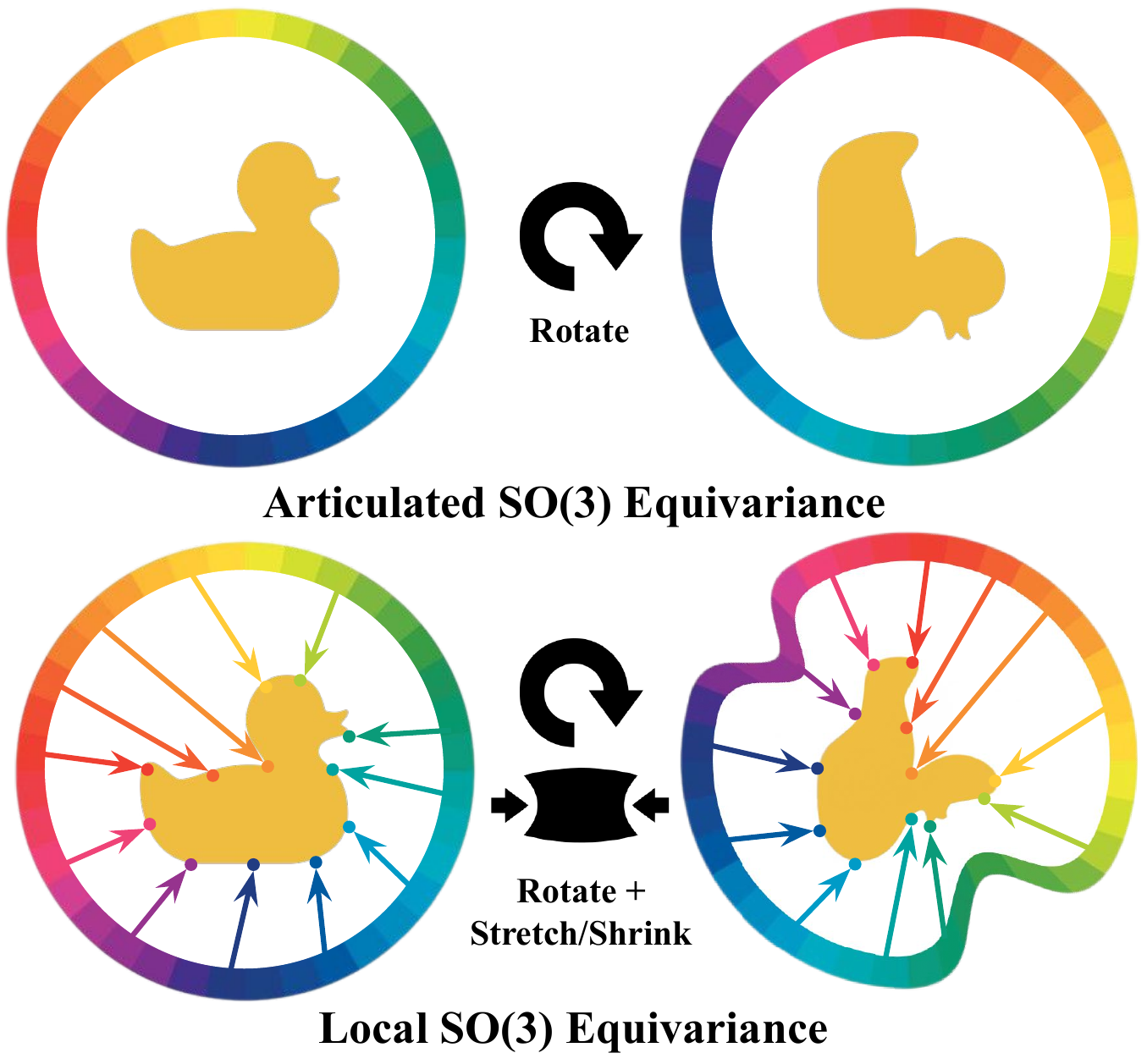}
    \caption{\textbf{SO(3) Equivariant Pose vs. Tightness.}  Rainbow circle is the feature $\mathcal{F}(\mathbf{X})$, for articulated SO(3)-equiv, $\mathcal{T}$ denotes approximate rigid transformation of body part, while for our case, where the clothing roughly deforms with human poses, it refers to the tightness vector rotation.}
    \label{fig:duck}
\end{figure}

Formally, we discretize $\mathrm{SO}(3)$ into a finite rotation group $\mathcal{G}$ with $|\mathcal{G}|=60$, where each element $\mathbf{g}_j$ corresponds to a rotation from the icosahedral group. As shown in \cref{fig:duck}, a continuous rotation amounts to a specific permutation of $\mathcal{G}$. Following prior work~\cite{feng2023generalizing}, we use this rotation group and its permutation operator to approximate the $\mathrm{SO}(3)$ rotation.

Given an input point cloud $\mathbf{X}$, the EPN~\cite{chen2021equivariant} network produces an $\mathrm{SO}(3)$-equivariant feature $\mathcal{F} \in \mathbb{R}^{N \times O \times C}$, where $N$ points and $O=60$ group elements in $\mathcal{G}=\{\mathbf{g}_1,\dots,\mathbf{g}_O\}$ each have a feature vector of size $C$. Mean pooling over the group dimension yields an invariant feature $\overline{\mathcal{F}} \in \mathbb{R}^{N \times C}$. These point-wise features capture localized, piecewise geometric information of the point cloud surface. In \arteq~\cite{feng2023generalizing}, they are aggregated by body parts to estimate articulated pose, whereas \modelname predicts tightness vectors mapping outer cloth to the inner body instead, as illustrated in~\cref{fig:duck}.

\subsection{Data Preparation}
\label{subsec:data}

The key step is to establish the dense correspondence mapping $\mathcal{\phi}: \mathbf{X} \rightarrow \mathbf{Y}$, between outer cloth points $\mathbf{X}$ and inner body points $\mathbf{Y}$. See mapping details in \textbf{``Anchor vs. Scatter Points''} at \cref{sec:implementation}.
Given each established pointwise correspondence $\mathcal{\phi}_i: \mathbf{x}_i \rightarrow \mathbf{y}_j$, $\mathbf{m}_k$ is the geodesic-closest marker of $\mathbf{y}_j$, all the other elements, including the tightness vector $\mathbf{v}_i$, label $l_i$, and confidence $c_i$, will be derived through:
\begin{equation}
\label{eq:elements}
    \begin{aligned}
    l_i &= k, \quad \mathbf{v}_i = \mathbf{y}_j - \mathbf{x}_i\\\
    c_i &= \exp(-\lambda \times g(\mathbf{m}_k,\mathbf{y}_j;\mathcal{S}_\mathbf{Y}))    
    \end{aligned}
\end{equation}
where $\lambda$ is the rate parameter of the exponential distribution to shrink it within the interval $[0,1)$, and  $g(\mathbf{m}_i, \mathbf{y}_j;\mathcal{S}_\mathbf{Y})$ is the geodesic distance defined on the inner surface $\mathcal{S}_\mathbf{Y}$.

\definecolor{equiv}{RGB}{251,164,68}
\definecolor{inv}{RGB}{112,180,143}

\begin{figure*}
    \centering
    \scriptsize
    \includegraphics[trim=000mm 000mm 000mm 000mm, clip=true, width=\linewidth]{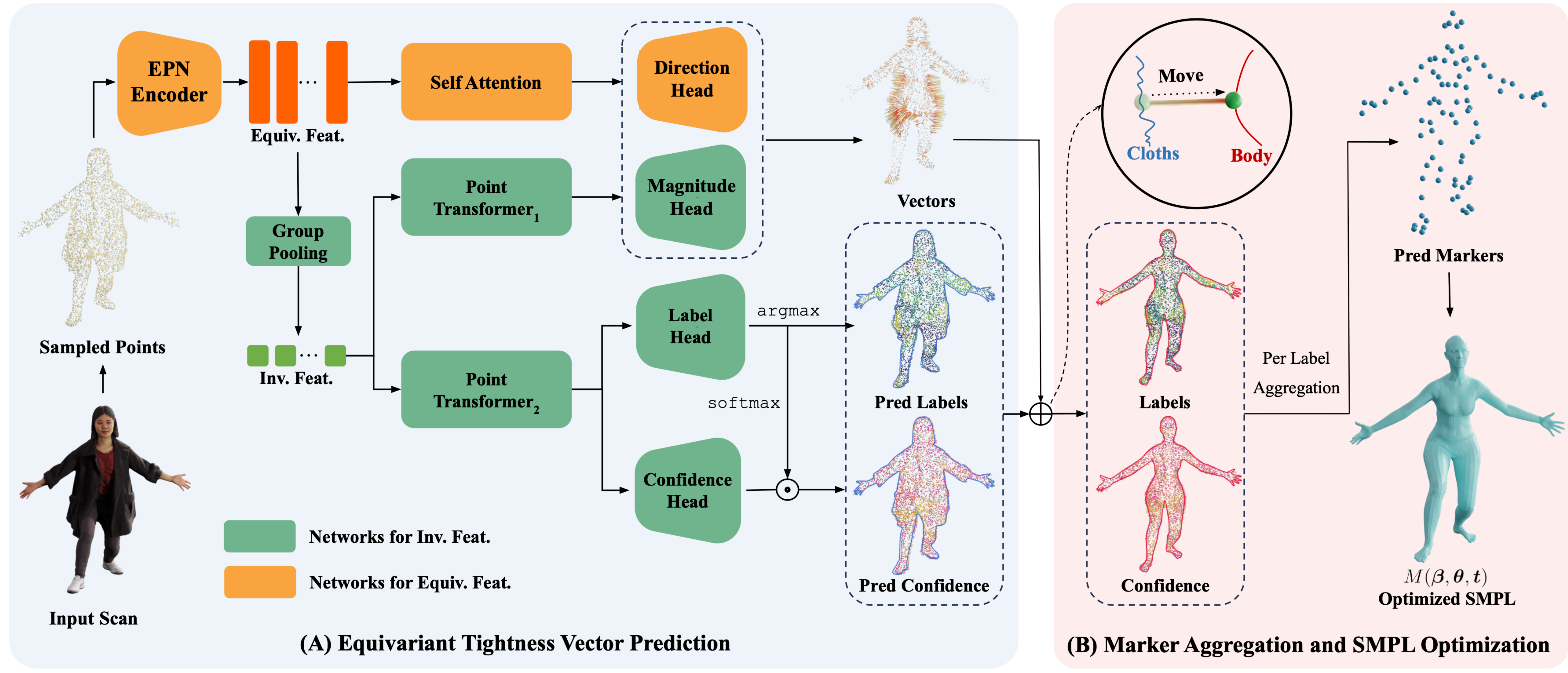}
    \caption{\textbf{\modelname Pipeline:} \textbf{1) Equivariant Tightness Vector Prediction}, which takes the sampled points $\mathbf{X}$ as input, and estimates the tightness directions $\mathbf{D}$ via \textcolor{equiv}{equivariant features $\mathbf{f}^\text{equiv}$ (\cref{subsec:vector})}, along with the tightness magnitudes $\mathbf{B}$, labels $\mathbf{L}$, and confidences $\mathbf{C}$ via \textcolor{inv}{invariant features $\mathbf{f}^\text{inv}$ (\cref{subsec:label-confidence})}. With these ingredients, in \textbf{2) Marker Aggregation and SMPL Optimization}, the points move inward along the tightness vectors, forming body-shaped point clouds. These points are weighted and aggregated (\cref{subsec:marker-smpl}), based on their labels and confidences, to produce final markers for SMPL fitting.}
    \label{fig:pipeline}
\end{figure*}

\subsection{Tightness Vector Prediction}
\label{subsec:vector}

``Tightness vector $\mathbf{v}_i$'' is the fundamental component of our system. Other components, including label $\mathbf{l}_i$ and confidence $\mathbf{c}_i$, are both derived from $\mathbf{v}_i$, as~\cref{eq:elements} shows. The tightness vector comprises two components: direction $\mathbf{d}_i$ and magnitude $b_i$, \ie $\mathbf{v}_i = b_i \mathbf{d}_i$. The direction highly correlates with human articulated poses, thus it is learned with local approximate SE(3) equivariance, ensuring the tightness vector consistently maps the cloth surface to the body surface. While the magnitude mainly reflects clothing displacements, highly correlating with clothing types and body regions; thus, it is learned from the invariance features, as illustrated in~\cref{fig:pipeline}.

\pheading{Direction Prediction.}
To obtain the direction $\mathbf{d}_i$ while preserving the equivariance feature $\mathbf{f}^\text{equiv}_i = \mathcal{F}_\text{EPN}(\mathbf{X})_i \in \mathbb{R}^{O \times C}$, we treat the pointwise features $\mathbf{f}^\text{equiv}_i$ independently, where $\mathcal{F}_\text{EPN}$ denotes the EPN network~\cite{chen2021equivariant}, and $O=60$ is the dimension of the rotation group $\mathcal{G}$. Specifically, we use a self-attention network $\mathcal{F}_\text{self-attn}$ to process the feature over the rotation group dimension ($O$), to ensure that each group element feature is associated with a group element $\mathbf{g}_j$ (which is a rotation in the discretized SO(3) group):
\begin{equation}
    \begin{aligned}
    \mathbf{f}^\text{equiv}_i &= \mathcal{F}_\text{EPN}(\mathbf{X})_i \\
    w_{ij} &= \mathcal{F}_\text{mlp}(\mathcal{F}_\text{self-attn}(\mathbf{f}^\text{equiv}_i)_j), \quad j \in \{0, 1, \dots, O-1\},\\
    \hat{\mathcal{R}}_i &= \mathcal{U} \begin{bmatrix} 1 & 0 & 0 \\ 0 & 1 & 0 \\ 0 & 0 & \det(\mathcal{U} \mathcal{V}^T) \end{bmatrix} \mathcal{V}^T, \quad \mathcal{U} \mathcal{D} \mathcal{V}^T = \sum_{j=1}^{O} w_{ij} \mathcal{R}_j. \\
    \end{aligned}
    \label{eq:mean_rotation}
\end{equation}
where $\mathcal{F}_\text{mlp}$ is the direction head parameterized with an MLP network, $\mathcal{R}_j$ is the rotation matrix for $\mathbf{g}_j$, and $\mathcal{D}$ is a diagonal matrix in the matrix decomposition. Finally, $\hat{\mathcal{R}}_i$ is derived, which is the rotation matrix for $\mathbf{x}_i$.
To transform from rotation to vector field, we multiply $\hat{\mathcal{R}}_i$ by a unit vector $\mathbf{v}_s$ (\eg $\mathbf{v}_s = \begin{bmatrix} 0 & 0 & 1 \end{bmatrix}$) to obtain the final $\hat{\mathbf{d}}_i$, \ie,
$\hat{\mathbf{d}}_i = \hat{\mathcal{R}}_i \mathbf{v}_s^\top$.

\pheading{Magnitude Prediction.}
To obtain $\hat{\mathbf{B}} = \{b_i \in \real\}_N$, we first aggregate (mean pooling) the feature over the rotation group dimension ($O$) for all points $\mathbf{X}$ to obtain the invariant feature $\mathbf{f}^\text{inv} = \overline{\mathcal{F}}_\text{EPN}(\mathbf{X}) \in \mathbb{R}^{N \times C}$. 
Instead of treating pointwise features $\mathbf{f}^\text{inv}_i = \overline{\mathcal{F}}_\text{EPN}(\mathbf{X})_i \in \real^{C}$ independently, we choose \emph{Point Transformer}~\cite{zhao2021point} $\mathcal{F}_\text{PT-1}$ to capture the contextual information outside this point. 

\begin{equation}
    \begin{aligned}
    \hat{\mathbf{B}} &= \mathcal{F}_\text{Mag}(\mathcal{F}_\text{PT-1}(\mathbf{f}^\text{inv}, \mathbf{X}; \delta)).  \\
    \end{aligned}
\end{equation}
where $\delta = \Theta(\mathbf{x}_i - \mathbf{x}_j)$ is the learned position embedding, parameterized with MLP $\Theta$, which encodes the relative positions between point pairs $\{\mathbf{x}_i, \mathbf{x}_j\}$, and $\mathcal{F}_\text{Mag}$ is the magnitude head, which is a MLP.

\subsection{Label and Confidence Prediction}
\label{subsec:label-confidence}
Along the estimated tightness vectors $\hat{\mathbf{V}}$, sampled points $\mathbf{X}$ shoot the inner body, where we will aggregate the predicted inner points, with their estimated marker label $\hat{\mathbf{L}}$ and confidence $\hat{\mathbf{C}}$, to obtain the final sparse markers $\hat{\mathbf{M}}$. Notably, here we also use $\textit{Point Transformer}$, denoted as $\mathcal{F}_\text{PT-2}$, since estimating labels and confidences is essentially a segmentation task, and the contextual information is crucial for distinguishing the symmetrical body parts (\ie, hands, feet).

\pheading{Label.}
Point Transformer $\mathcal{F}_\text{PT-2}$ and $\mathcal{F}_\text{Label}$ takes $\overline{\mathcal{F}}_\text{EPN}(\mathbf{X})  \in \mathbb{R}^{N \times C}$ with position $\mathbf{X} \in \mathbb{R}^{N \times 3}$ as input and outputs $\mathcal{P} \in \mathbb{R}^{N \times K}$, here $K=86$, with $\mathcal{P}(\mathbf{x}_i, \mathbf{m}_k)$ representing the probability of point $\mathbf{x}_i$ belonging to marker $\mathbf{m}_k$:

\begin{equation}
    \begin{aligned}
     \mathcal{P}(\mathbf{X}) &= \texttt{softmax}(\mathcal{F}_\text{Label}(\mathcal{F}_\text{PT-2}(\mathbf{f}^\text{inv}, \mathbf{X}; \delta))),\\
     \hat{\mathbf{L}} &= \texttt{argmax}(\mathcal{F}_\text{Label}(\mathcal{F}_\text{PT-2}(\mathbf{f}^\text{inv}, \mathbf{X}; \delta))).
     \end{aligned}
\end{equation}

\pheading{Confidence.}
Following LoopReg~\cite{bhatnagar2020loopreg}, we use Group Convolution $\mathbf{C}_k$ for each marker $\mathbf{m}_k$ , and soft aggregation to compute the confidence of each tightness vector:

\begin{equation}
\begin{aligned}
    \beta_{ik} &= \mathcal{F}_\text{Conf}(\mathcal{F}_\text{PT-2}(\mathbf{f}^\text{inv}, \mathbf{X}; \delta))_i \ast \mathbf{C}_k,\\
    \hat{c}_i &= \sum_{k=1}^{K} \mathcal{P}(\mathbf{x}_i, \mathbf{m}_k) \beta_{ik}.
\end{aligned}
\end{equation}
where both $\mathcal{F}_\text{Conf}$ and $\mathcal{F}_\text{Label}$ are MLPs, $\mathcal{P}(\mathbf{X}) \in \mathbb{R}^{N \times K}$.

Now we have all the ingredients for cooking, the final loss $\mathcal{L}$ is formulated as follows:

\begin{equation}
\label{eq:losses}
    \begin{gathered}
        \mathcal{L} = w_d \mathcal{L}_d + w_b \mathcal{L}_b + w_l \mathcal{L}_l + w_c \mathcal{L}_c,\\
        \mathcal{L}_d = \sum_{i=1}^{N} \frac{\hat{\mathbf{v}}_i \cdot \mathbf{v}_i}{\|\hat{\mathbf{v}}_i\| \|\mathbf{v}_i\|}, \quad \mathcal{L}_b = \frac{1}{N} \sum_{i=1}^{N} (\hat{\mathbf{b}}_i - \mathbf{b}_i)^2,\\
        \mathcal{L}_l = - \frac{1}{N} \sum_{i=1}^{N} \log(\mathcal{P}(\mathbf{x}_i, \mathbf{m}_{k=l_{i}})),
         \quad \mathcal{L}_c = \frac{1}{N} \sum_{i=1}^{N} (\hat{\mathbf{c}}_i - \mathbf{c}_i)^2.\\
    \end{gathered}
\end{equation}
where $w_d, w_b, w_l, w_c$ separately represents the weighted factor of losses of direction, magnitude, label and confidence, see more technical details in \cref{sec:implementation}. 

\pheading{End2End Training.} 
\cref{eq:losses} indicates that the training process adopts multitask supervision rather than end-to-end training. Considering an end-to-end system is valuable, but there are practical issues: 1) Top-m selection isn't differentiable, and top-all aggregation is time-consuming. 2) End-to-end training may lead to suboptimal shortcuts, affecting useful intermediate output for downstream tasks like garment parsing. Thus, multitask supervision remains essential, especially with limited data.

\subsection{Marker Aggregation and SMPL Regression}
\label{subsec:marker-smpl}
\pheading{Marker Aggregation.}
Along the predicted tightness vector $\hat{\mathbf{v}}_i$, the outer points $\mathbf{x}_i$ are shot towards inner points $\hat{\mathbf{y}}_i$, via $\hat{\mathbf{y}}_i = \mathbf{x}_i + \hat{\mathbf{v}}_i$. Then the body markers are obtained through weighted aggregation as follows:

\begin{equation}
    \hat{\mathbf{m}}_k = \frac{\sum_{i=1}^{m} \hat{\mathbf{y}}_i^{\hat{l}_i = k} \cdot (\hat{c}_i^{\hat{l}_i=k})^\alpha}{\sum_{i=1}^{m} (\hat{c}_i^{\hat{l}_i=k})^\alpha}
\end{equation}
where only top-$m$ points with the highest confidences among those where the $\hat{l}_i$ coincides with k are aggregated, $\alpha$ further amplifies their influence, enhancing results in practice. With these aggregated markers, SMPL parameters are optimized: 

\begin{equation}
    \min_{\boldsymbol{\theta}, \boldsymbol{\beta}, \mathbf{t}} \sum_{k=1}^{K} \left\| \tilde{\mathbf{m}}_k - \hat{\mathbf{m}}_k \right\|^2
\end{equation}
, where $\tilde{m}_k$ denotes the markers on current SMPL estimate. Specifically, we use a damped Gauss-Newton optimizer based on the Levenberg–Marquardt algorithm~\cite{roweis1996levenberg}. 

We initially considered a Geodesic-based weighted sum but opted for a Euclidean approach due to key limitations: \textbf{1) Surface:} The computation of geodesic distance requires a surface, while \modelname uses raw point clouds common in human capture. \textbf{2) Simplicity:} Surface construction (\eg Poisson reconstruction) and geodesic computation are time-consuming. 
\textbf{3) Minor Benefits:} We only select top-3 confidence points for marker aggregation, forming a tiny cluster near the body surface. The Euclidean assumption yeilds a marker error of just 0.31\% relative to body scale. Due to minimal improvement vs. high computational cost, we chose the Euclidean-based option, which already achieves SOTA.

\section{Experiments}
\label{sec:experiments}

\subsection{Datasets.}
\label{sec:datasets}
We select CAPE~\cite{ma2020cape} and \ddress~\cite{wang20244ddress}, both featuring pre-captured underlying bodies, and a wide range of human poses, shapes, and clothing variations. Unlike \ddress, CAPE has tighter-fitting clothing and less dynamics.
Specifically, CAPE~\cite{ma2020cape} contains 15 subjects with different body shapes; we split them as 4:1 to evaluate the robustness against ``body shape and garment variations''. Following NICP~\cite{marin24nicp}, we subsample by factors of 5 and 20 for training and validation sets, resulting in 26,004 training frames and 1,021 validation frames.
For 4D-Dress~\cite{wang20244ddress}, which contains 32 subjects with 64 outfits across over 520 motion sequences, we use the official split which selects 2 sequences per outfit to evaluate the robustness against ``body pose and clothing dynamics variations''. After subsampling by factors of 1 and 10 for training and validation respectively, we obtain 59,395 training frames and 1,943 validation frames. All the models are trained and tested with the above setting.

\subsection{Metrics.}
\label{sec:metrics}

\qheading{Distance Metrics.} Three metrics are included: 1) vertex-to-vertex (V2V) distance in cm, between the correspondence vertices of our fitted and provided \gt SMPL bodies, 2) joint position error (MPJPE) in cm, measuring the Euclidean error of $J$ SMPL joints, and 3) bidirectional Chamfer Distance in cm, between the predicted inner body points (w/o SMPL fitting) and \gt SMPL bodies, a metric specifically designed to evaluate tightness-aware methods. \Cref{tab:benchmark,tab:ablation,tab:chamfer} uses these distance metrics, with lower values indicating more accurate body fits.

\pheading{Directional Metrics.}\label{para:metrics} We ablate the effectiveness of Equivariance features by measuring the correctness of tightness vector direction $\mathbf{d}_i$, as this is the only element regressed from equivariance features. We calculate the angular error between the estimated and \gt tightness directions (detailed in~\cref{subsec:data}). Specifically, in~\cref{tab:equiv}, we report both the mean and median cosine distance, with lower numbers indicating more accurate tightness directions. 

\subsection{Technical Details}
\label{sec:implementation}

\qheading{Implementation.}
We train \modelname using the Adam optimizer with a learning rate of 1e-4 and a batch size of 2. The training process requires 21 epochs on \ddress and 39 epochs on CAPE, taking about 4 days on a single NVIDIA GeForce RTX 4090. The number of sampled points in scans is 5000. We set all loss weights \(w_d, w_b, w_l, w_c\) to 1.0. For the EPN network used in \cref{subsec:vector}, the radius is set to 0.4, and the number of layers is 2. The marker-fitting ($K=86$) process converges in approximately 5 seconds after 80 steps per subject. Following \cite{wang2021ptf}, we adopt a two-stage optimization: first optimizing the $\shapecoeff[:2]$ and $\posecoeff$ for 30 steps with $\text{lr}=5e-1$, then optimizing all parameters for 50 steps with $\text{lr}=2e-1$. 

\pheading{Anchor Points vs. Scattered Points.}
It should be noted that all the points are uniformly sampled from the \emph{surface} instead of \emph{vertices}. $\mathbf{Y}$ is sampled from the SMPL mesh $M(\shapecoeff, \posecoeff, \mathbf{t})$. These points $\mathbf{y}_j$ are shot along their normals to intersect the outer surface, termed \emph{Anchor Points} $\mathbf{x}_j$. Then we uniformly sample points $\mathbf{x}_i \in \mathbf{X}$, termed \emph{Scattered Points}. Each scattered point finds its closest anchor point based on geodesic distance $g(\mathbf{x}_i, \mathbf{x}_j;\mathcal{S}_\mathbf{X})$, $\mathcal{S}_\mathbf{X}$ is the outer clothed triangle mesh surface; if this distance is below a threshold (= 0.01), it shares the corresponding inner point $\mathbf{y}_j$ and its corresponding marker $\mathbf{m}_k$. Otherwise, the closest inner point, based on Euclidean distance, is selected instead.

\subsection{Quantitative Results.}
\label{sec:quantitative}

\begin{table}[t]
\setlength{\tabcolsep}{5pt}
\centering{
\scriptsize
  \begin{tabular}{c|ccc|ccc}
    Methods & \multicolumn{3}{c|}{CAPE~\cite{ma2020cape}} & \multicolumn{3}{c}{\ddress~\cite{wang20244ddress}}\\
     &  V2V $\downarrow$ & MPJPE $\downarrow$ & CD $\downarrow$ & V2V $\downarrow$ & MPJPE $\downarrow$ & CD $\downarrow$ \\
     \shline
     \multicolumn{7}{c}{Tightness-agnostic} \\
     \hline
     NICP~\cite{marin24nicp} & 1.726 & 1.343 & - & 4.754 & 3.654 & - \\
     \arteq~\cite{feng2023generalizing} & 2.200 & 1.557 & - & 2.328 & 1.657 & - \\
     \hline
     \multicolumn{7}{c}{Tightness-aware} \\
     \hline
     \ipnet~\cite{bhatnagar2020ipnet} & 2.593 & 1.917 & 1.110 & 3.826 & 2.625 & 1.262 \\
     PTF~\cite{wang2021ptf} & 2.036 & 1.497 & 1.219 & 2.796 & 2.053 & 1.239 \\
     Ours & \textbf{1.647} & \textbf{0.922} & \textbf{1.019}  & \textbf{1.939}  & \textbf{1.116} & \textbf{1.065} \\
  \end{tabular}
}
\caption{
    \textbf{Quantitative Comparison with SOTAs.} \modelname clearly outperforms SOTAs, whether tightness-agnostic or -aware, in both CAPE and \ddress across all metrics. In \ddress-MPJPE, it surpasses the \arteq by nearly \textcolor{ForestGreen}{$32.6\%$}. Notably, for a fair comparison, no post-refinement is introduced to NICP~\cite{marin24nicp} here, see NICP w/ post-refinement at \cref{tab:chamfer}.
}
\label{tab:benchmark}
\end{table}

\begin{table}[t]
\setlength{\tabcolsep}{5pt}
\centering{
\resizebox{\linewidth}{!}{
\scriptsize
  \begin{tabular}{c|ccc|ccc|ccc}
    Methods & \multicolumn{3}{c|}{Loose Garment} & \multicolumn{3}{c|}{Extreme Shape} & \multicolumn{3}{c}{Challenging Pose}\\
     &  V2V $\downarrow$ & MPJPE $\downarrow$ & CD $\downarrow$ & V2V $\downarrow$ & MPJPE $\downarrow$ & CD $\downarrow$ & V2V $\downarrow$ & MPJPE $\downarrow$ & CD $\downarrow$\\
     \shline
     \multicolumn{10}{c}{Tightness-agnostic} \\
     \hline
     NICP & 9.113 & 6.940 & - & 3.906 & 3.165 & - & 4.523 & 3.543 & -\\
     \arteq & 3.428 & 2.746 & - & 2.137 & 1.460 & - & 2.420 & 1.741 & - \\
     \hline
     \multicolumn{10}{c}{Tightness-aware} \\
     \hline
     \ipnet & 4.441 & 2.932 & 1.652 & 3.751 & 3.002 & 1.240 & 3.860 & 2.747 & 1.273\\
     PTF & 3.264 & 2.341 & 1.600 & 3.122 & 2.645 & 1.216 & 2.914 & 2.221 & 1.249\\
     Ours & \textbf{2.276} & \textbf{1.455} & \textbf{1.340} & \textbf{1.831}  & \textbf{1.074} & \textbf{0.998} & \textbf{1.992} & \textbf{1.171} & \textbf{1.070}\\    
  \end{tabular}
  }
}
\caption{
    \textbf{Challenge Subsets of \ddress.} For further evaluation of challenging cases, we test all the models on challenging subsets of 4D-Dress, including Loose Garment, Extreme Shape and Challenging Pose. See~\cref{sec:quantitative} for the details on how the challenging subsets of \ddress are filtered. ETCH still leads on ALL these challenging subsets, each corresponding to a specific challenging aspect, thereby its superior OOD generalization improves the overall performance.
}
\label{tab:challenge_subsets}
\end{table}

We compare our method with multiple SOTA baselines, spanning from tightness-aware approaches, \ie, \ipnet~\cite{bhatnagar2020ipnet} and PTF~\cite{wang2021ptf}, to tightness-agnostic ones, \ie, NICP~\cite{marin24nicp} and \arteq~\cite{feng2023generalizing}, as shown in \cref{tab:benchmark}. 

\pheading{Baselines.} \modelname achieves superior performance across all datasets and metrics. In particular, on CAPE, among all the competitors, our approach reduces the V2V error by \textcolor{ForestGreen}{$4.6\% \sim 36.5\%$} and MPJPE by \textcolor{ForestGreen}{$31.3\% \sim 51.9\%$}; on 4D-Dress, the improvement is even more significant with a \textcolor{ForestGreen}{$16.7\% \sim 59.2\%$} decrease in V2V error and \textcolor{ForestGreen}{$32.6\% \sim 69.5\%$} in MPJPE. 
Among tightness-aware methods (\ie, \ipnet, PTF and Ours), under bidirectional Chamfer Distance, our method achieves \textcolor{ForestGreen}{$8.2\% \sim 16.4\%$} improvement on CAPE and \textcolor{ForestGreen}{$14.0\% \sim 15.6\%$} on \ddress between the predicted inner points/meshes (w/o SMPL fitting) and \gt SMPL bodies. 
We attribute this leading edge to our innovative design of ``tightness vector'', which effectively disentangles clothing and body layers, simplifying the fitting of loosely clothed humans into a bare-body fitting problem. This makes our method more robust to variations of bodies (shapes, poses) and clothing (types, fitness, dynamics). 

It is worth noting that among optimization-based methods (\ie, NICP, \ipnet, and PTF), additional priors, regularizers, or post-refinement are incorporated during SMPL optimization, such as pose priors (\eg, VPoser~\cite{SMPLX:2019}, GMM~\cite{federica2016smplify}), shape regularizer to encourage $\shapecoeff$ to be small, inner-to-body point-mesh distance regularizer (\ie, PTF, \ipnet), and ``Chamfer-based Post-Refinement'' of NICP $\dagger$ in \cref{tab:chamfer}. 
In contrast, our method optimizes SMPL using only the 86 markers (\cref{subsec:marker-smpl}) directly predicted by \modelname as optimization targets, \textit{\textbf{without bells and whistles}} (\ie, pose priors, and geometric constraints), yet still achieves substantial performance gains, further validating our key design. In addition, the body fitting error could arise from either pose or shape errors; joint error MPJPE in~\cref{tab:benchmark} only confirms pose accuracy, so we also evaluate ``shape accuracy'' in \cref{fig:consistency}, showing average \textcolor{ForestGreen}{$49.9\%$} improvement compared to other optimization-based approaches.

\pheading{Challenge Subsets.}
As \cref{sec:datasets} explains, our train-test split (in ~\cref{tab:benchmark}) of CAPE is across subjects (unseen shapes and outfits), and \ddress is along the motion sequence (unseen poses). For further evaluation of challenging cases, we test all the models on challenging subsets of \ddress. Specifically, in~\cref{tab:challenge_subsets}, we split three challenging subsets (top 10\% outliers) from the test set of \ddress, including \textit{Loose Garments} (thickest outfits measured by tightness magnitude $b_i$), \textit{Extreme Shapes} (most outlier shapes measured by SMPL shape params $\shapecoeff[:3]$), and \textit{Challenging Poses} (most outlier poses measured by SMPL pose params $\posecoeff$). \modelname still leads on ALL these challenging subsets, thus its superior OOD generalization improves the overall performance.

\subsection{Qualitative Results.}
\label{sec:qualitative}

\begin{figure*}[p]
    \centering
    \vspace{-3.0em}
    \scriptsize
    \includegraphics[trim=000mm 000mm 000mm 000mm, clip=true, width=\linewidth]{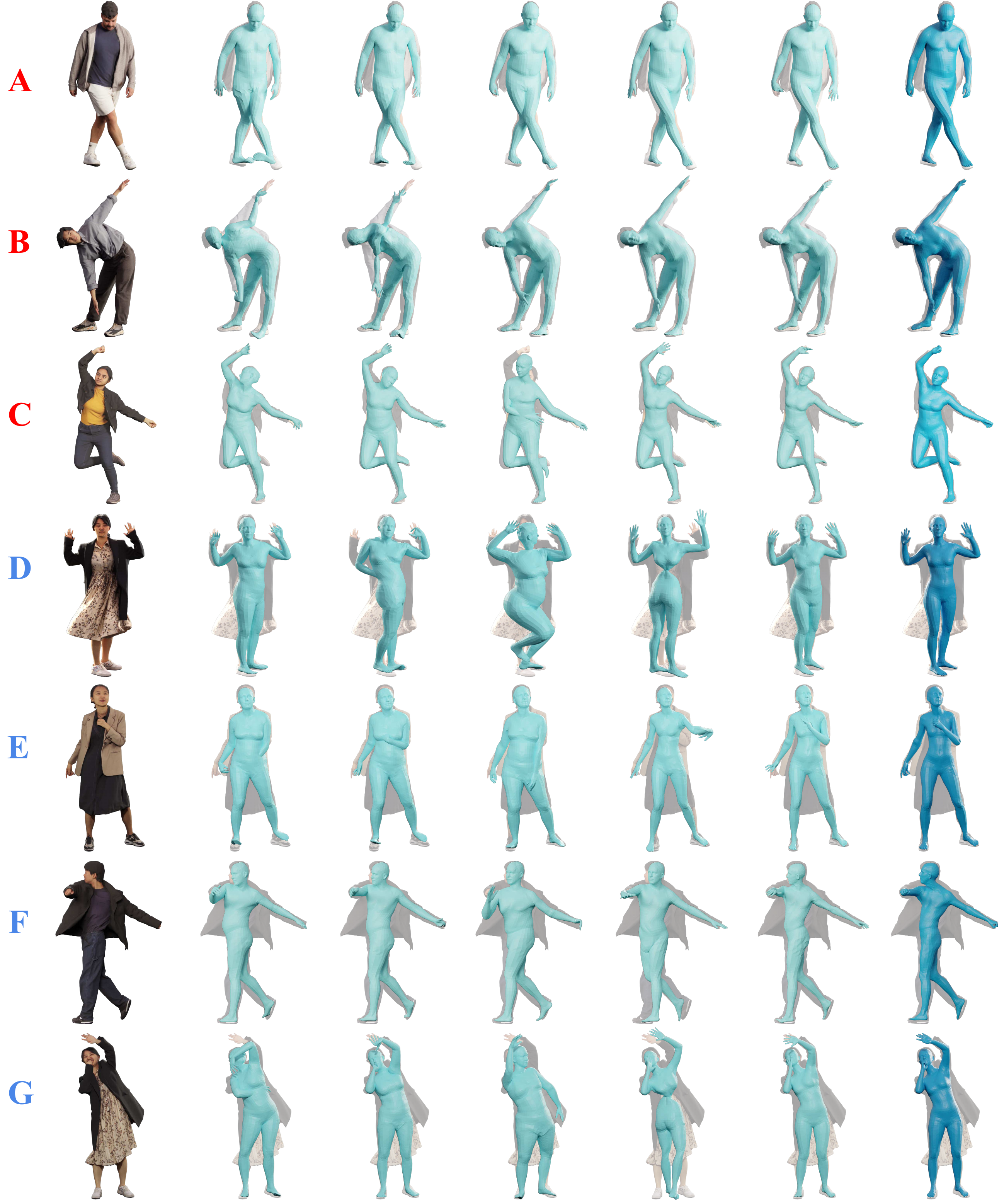}
    \begin{tabularx}{\linewidth}{
        @{}p{0.053\linewidth}@{}
        >{\centering\arraybackslash}p{0.116\linewidth}
        >{\centering\arraybackslash}p{0.116\linewidth}
        >{\centering\arraybackslash}p{0.116\linewidth}
        >{\centering\arraybackslash}p{0.116\linewidth}
        >{\centering\arraybackslash}p{0.116\linewidth}
        >{\centering\arraybackslash}p{0.116\linewidth}
        >{\centering\arraybackslash}p{0.090\linewidth}} 
        & Scan & \ipnet~\cite{bhatnagar2020ipnet} & PTF~\cite{wang2021ptf} & NICP~\cite{marin24nicp} & \arteq~\cite{feng2023generalizing} & Ours & GT\\
    \end{tabularx}
    \caption{\textbf{Comparison on \textcolor{Red}{Challenging Poses (A, B, C)} and \textcolor{NavyBlue}{Hard Garments (D, E, F, G)}.} (A) Crossed Legs Pose; (B) Extended Triangle Pose; (C) Asymmetric Limb Pose; (D) Dress Twist; (E) Open Blazer; (F) Flowing Puffer; (G) Waving in Dress. Our method consistently achieves superior pose and shape alignment with \gt SMPL. While both \arteq~\cite{feng2023generalizing} and ours appear robust to challenging poses -- in case A, others misplace the left and right legs; in case B, they misrotate the torso or the head; in case C, the head and the legs are unaligned with \gt SMPL -- our SMPL results are still better than \arteq's, particularly in cases A (lower legs), B (raised forearm) and C (left forearm and abdomen). Our advantages are more clear for loose garments. In cases D and F, involving loose clothing and torso rotation, other methods mispredict head or hip rotations, with \arteq showing a ``Taffy and Bowtie'' distortion. In case E, they incorrectly predict the left arm position, while in case G, they misplace limbs.}
    \label{fig:comparison}

\end{figure*}

\begin{figure}[t]

    \centering{
    \includegraphics[trim=000mm 000mm 000mm 000mm, clip=true, width=\linewidth]{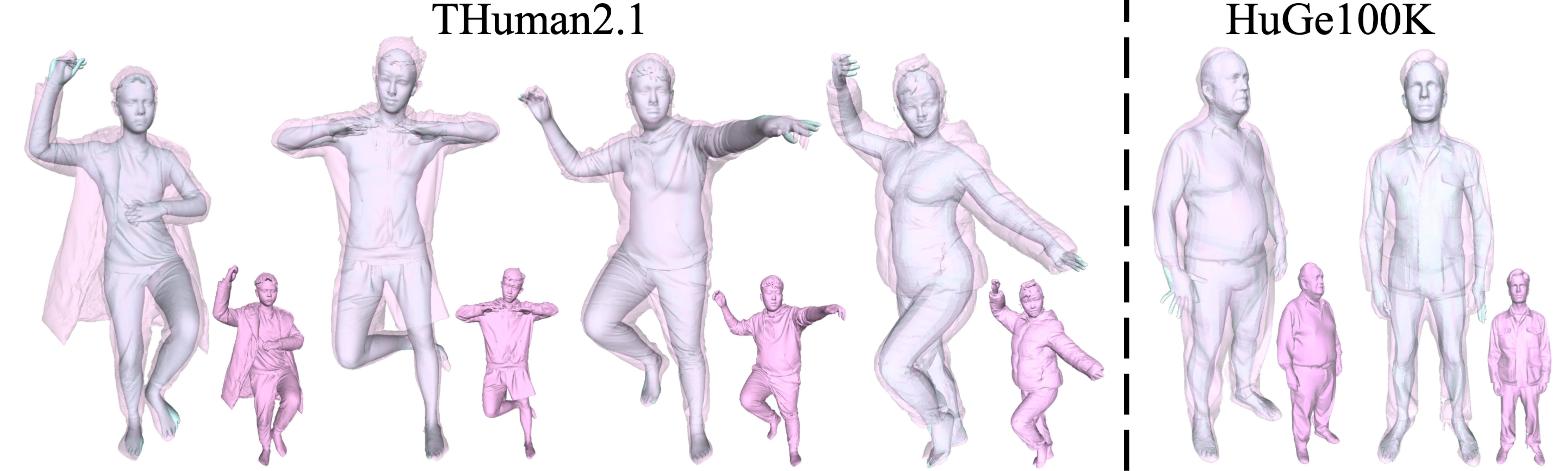}

    \caption{\textbf{Cross Dataset Testing (Unseen Garments, Shapes, Poses).} ETCH trained on 4D-Dress can generalize well on unseen THuman2.1 scans and HuGe100K generated 3D humans~\cite{zhuang2024idolinstant}
    }
    \label{fig:crossdataset}}

\end{figure}

In \cref{fig:comparison}, each row illustrates a challenging case: A, B and C focus on difficult poses, while D, E, F and G highlight loose garments with large dynamic deformations. Our method achieves pixel-level body alignment, showing its superior robustness to various poses, shapes, and garments. More analysis is detailed in the caption of~\cref{fig:comparison}. \cref{fig:crossdataset} shows ETCH trained on \ddress generalize well on unseen THuman2.1 scans and HuGe100K generated 3D humans~\cite{zhuang2024idolinstant}. Please check out more video results in \web.

\begin{table*}[t]
\scriptsize
\centering{
  \begin{tabular}{c|cc|cc|ccc||cc|cc}
     & \multicolumn{7}{c||}{Ablation Settings} & \multicolumn{2}{c|}{CAPE~\cite{ma2020cape}} & \multicolumn{2}{c}{\ddress~\cite{wang20244ddress}}\\
    \hline
     & \multicolumn{2}{c|}{Tightness} & \multicolumn{2}{c|}{Correspondence} & \multicolumn{3}{c||}{Features for Direction $\mathbf{d}_i$} & \multirow{2}{*}{V2V $\downarrow$} & \multirow{2}{*}{MPJPE $\downarrow$} & \multirow{2}{*}{V2V $\downarrow$} & \multirow{2}{*}{MPJPE $\downarrow$}\\
     \cline{2-8}
     Settings & direction & scalar & dense & markers & Inv & XYZ & Equiv & & & & \\
     \shline
    Ours & \cmark & \cmark & \xmark & \cmark & \xmark & \xmark & \cmark & \textbf{1.647} & \textbf{0.922} & \textbf{1.939} & \textbf{1.116} \\
    \hline
    A. & \cmark & \cmark & \xmark & \cmark & \xmark & \cmark & \xmark & 1.661 & 0.925 & 2.033 & 1.134 \\
    B. & \cmark & \cmark & \xmark & \cmark & \cmark & \cmark & \xmark & 1.663 & 0.926 & 2.307 & 1.314 \\
    C. & \cmark & \cmark & \cmark & \xmark & \xmark & \xmark & \cmark & 1.909  & 1.451  & 2.285  & 1.466  \\
    D. & \xmark & \cmark & \cmark & \xmark & \xmark & \xmark & \cmark & 1.777  & 1.342  & 2.410  & 1.608  \\
    E. & \cmark & \cmark & \xmark & \cmark & \cmark & \xmark & \xmark & 
    1.888 & 1.101 & 2.842 & 2.014 \\
  \end{tabular}}
\caption{
    \textbf{Ablation Study of \modelname.} Please check \cref{sec:ablation} for more in-depth analysis, and \cref{tab:equiv,fig:equiv} to explore OOD generalization of equivariance features. For simplicity, ``Inv'' denotes Invariance Features, ``Equiv'' denotes Equivariance Features, ``XYZ'' denotes XYZ-Positions. The full-featured \modelname is referred to as ``Ours'', while variants are labeled ``Ours-X''. Ours-A and Ours-B replace equivariance features with xyz-positions and/or invariance features, while Ours-E only uses invariance features. Although invariant feature is inherently incompatible with Direction prediction, we add Ours-E here for more comprehensive comparison. Ours-C and Ours-D use dense correspondence, with Ours-D removing the direction term to assess its necessity. 
}
\label{tab:ablation}
\end{table*}

\subsection{Ablation Studies}
\label{sec:ablation}

\begin{figure}[t]
    \centering
    \scriptsize
    \includegraphics[trim=000mm 000mm 000mm 000mm, clip=true, width=\linewidth]{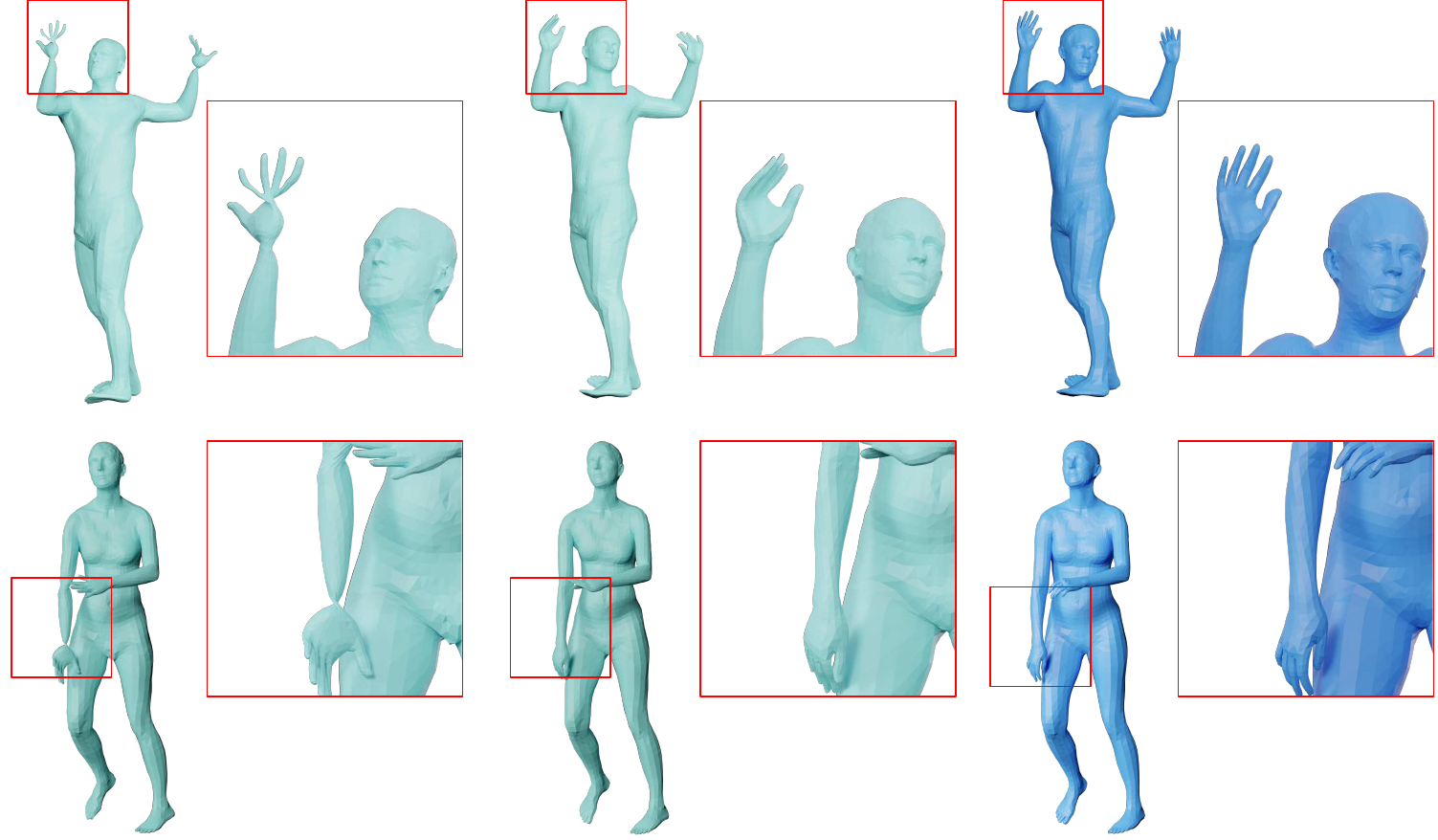}
    \begin{tabularx}{\linewidth}{
        >{\centering\arraybackslash}X
        >{\centering\arraybackslash}X
        >{\centering\arraybackslash}X} 
        Ours-C & Ours & GT body \\
    \end{tabularx}
    \caption{\textbf{Sparse Markers vs. Dense Correspondence.} Incorrect rotation angles and flipped configurations are evident in the head and hand regions of the first row, and in the forearm and hand regions of the second row. More quantitative results in~\cref{tab:ablation} and its analysis in~\cref{sec:ablation}.
    }
    \label{fig:marker_corr}
\end{figure}

\begin{figure*}[t]
    \scriptsize
    \centering{
    \includegraphics[trim=000mm 000mm 000mm 000mm, clip=true, width=0.9\linewidth]{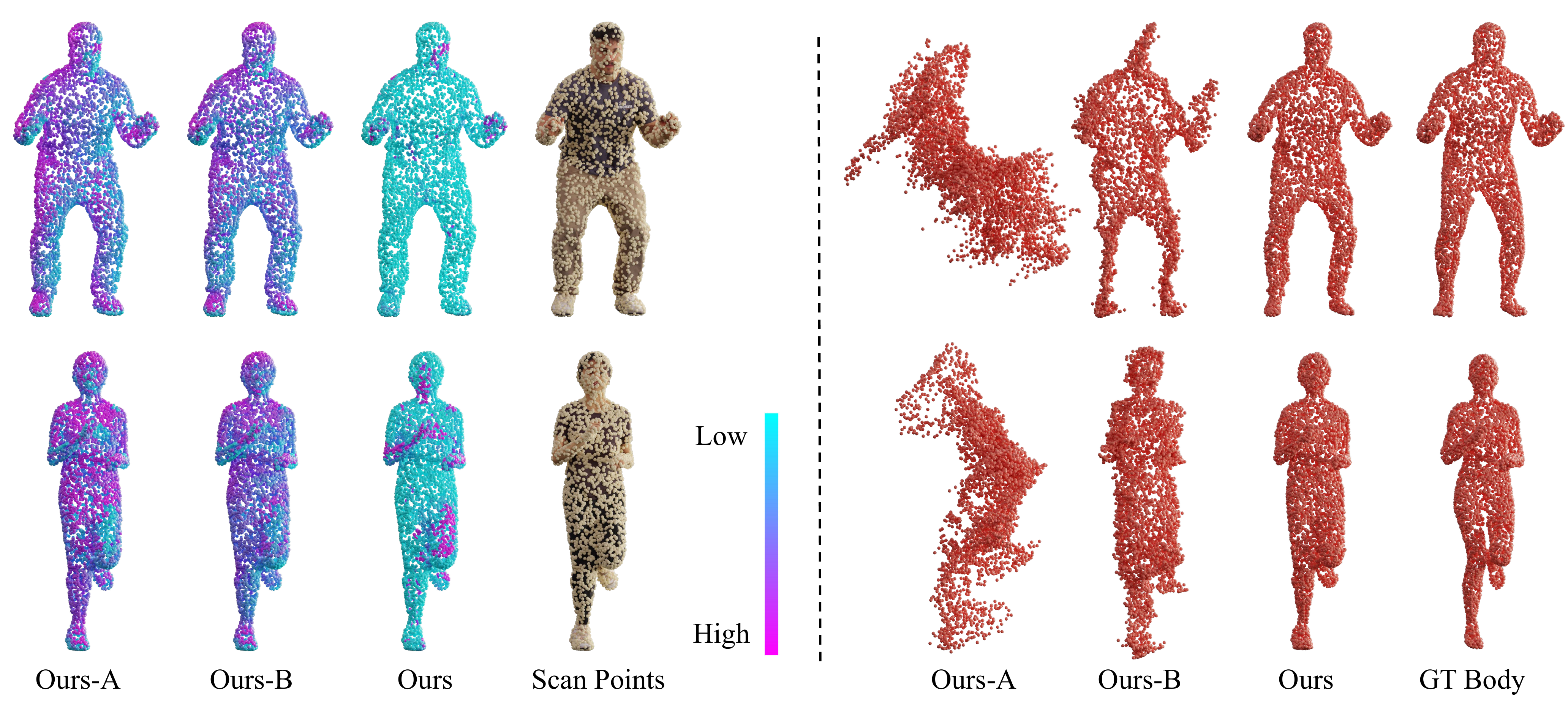}
    \caption{\textbf{Equivariance for OOD.} We test multiple variants of \modelname on one-shot settings to illustrate the out-of-distribution (OOD) generalization of equivariant features, by replacing the equivariant features (Ours) with xyz-positions (Ours-A) or xyz-positions+invariance (Ours-B). The visualization shows scan points colored by angular error (left) and predicted inner points (right). The \gradientRGB{error bar}{82,216,217}{195,69,206} indicates cosine error of tightness vector. 
    }
    \label{fig:equiv}}
\end{figure*}

\begin{figure}[t]
    \centering
    \includegraphics[trim=000mm 000mm 000mm 000mm, clip=true, width=1.0\linewidth]{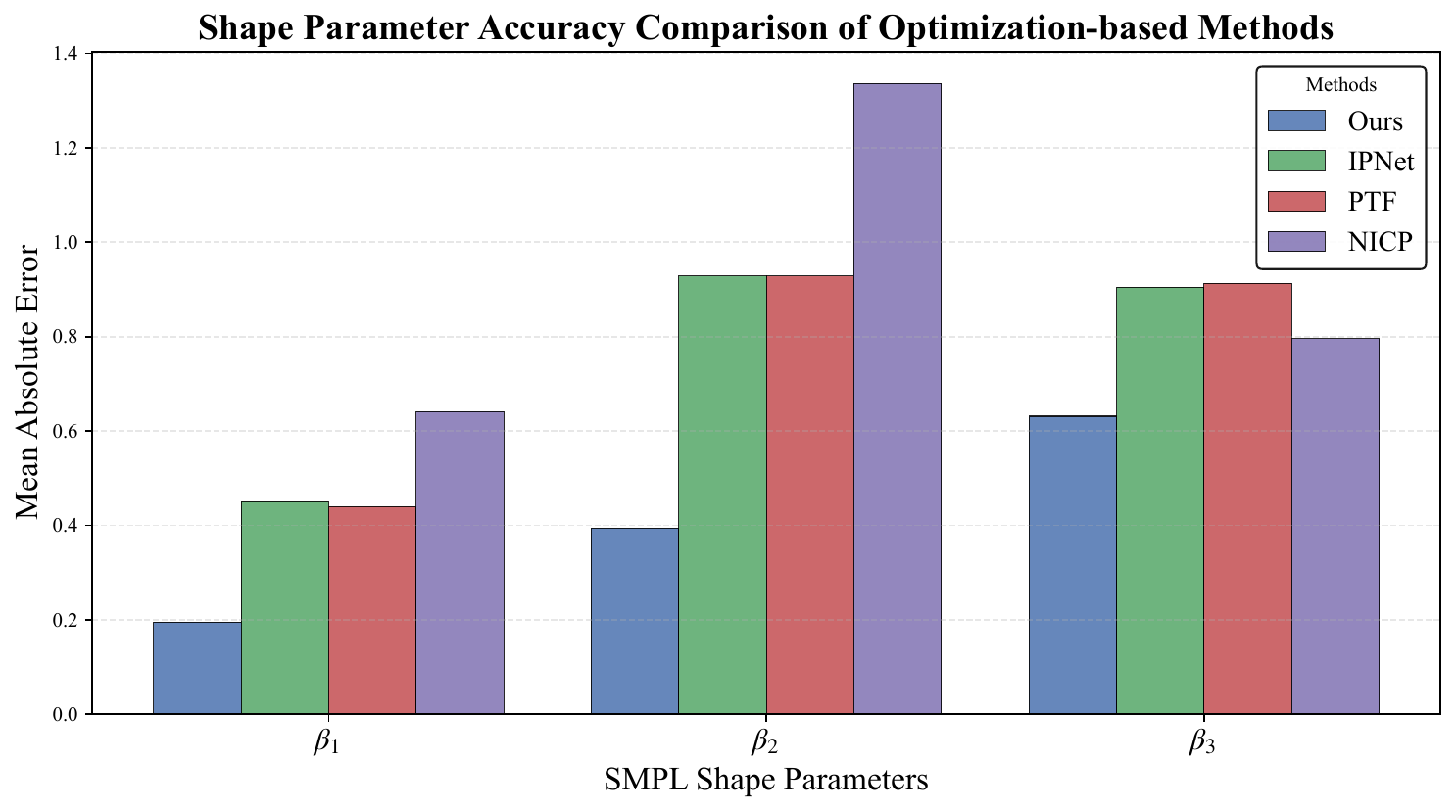}
    \caption{\textbf{Shape Accuracy Analysis.} We calculate the Mean Absolute Error (MAE) on \ddress for the first three principal shape parameters, \(\shapecoeff_{[:3]} \in \mathbb{R}^{3}\). Our method shows a significant advantage in shape accuracy, with an average improvement of \textcolor{ForestGreen}{$49.9\%$}.}
    \label{fig:consistency}
\end{figure}

We conduct ablation studies to explore strategies like ``Sparse Marker vs. Dense Correspondence'' (\cref{tab:ablation,fig:marker_corr}) and validate design choices, such as the Tightness Scalar in ``Tightness Vector vs. Scalar'' (\cref{tab:ablation}), Equivariance features in ``w/ Equiv vs. w/o Equiv'' under full training set and one-shot settings (\cref{tab:equiv,fig:equiv}), the influence of chamfer-based post-reinforcement (\cref{tab:chamfer}), and shape accuracy (\cref{fig:consistency}). The full-featured \modelname is referred to as ``Ours'', while variants are labeled ``Ours-X''. Ours-A and Ours-B replace equivariance features with xyz-positions and/or invariance features, Ours-E only uses invariance features. Although invariance feature is inherently incompatible with Direction prediction, we add Ours-E here for more comprehensive comparison. Ours-C and Ours-D use dense correspondence, with Ours-D removing the direction term to assess its necessity. 

\begin{table}[t]
\scriptsize
\centering{
\setlength{\tabcolsep}{4pt}
  \begin{tabular}{c|c|c|c||cc|cc}
     & \multicolumn{3}{c|}{Features for Direction $\mathbf{d}_i$} & \multicolumn{2}{c|}{CAPE~\cite{ma2020cape}} & \multicolumn{2}{c}{\ddress~\cite{wang20244ddress}}\\
    \cline{2-4}
    Settings & Inv & XYZ & Equiv & Mean $\downarrow$& Median $\downarrow$ & Mean $\downarrow$ & Median $\downarrow$\\
     \shline
    \multirow{1}{*}{Ours} & \multirow{1}{*}{\xmark} & \multirow{1}{*}{\xmark} & \multirow{1}{*}{\cmark} & \textbf{0.616} & \textbf{0.0535} & \textbf{0.919} & \textbf{0.315}  \\
    \hline
    \multirow{1}{*}{A.} & \multirow{1}{*}{\xmark} & \multirow{1}{*}{\cmark} & \multirow{1}{*}{\xmark} & 0.763 & 0.527 & 0.986 & 0.988  \\
    \multirow{1}{*}{B.} & \multirow{1}{*}{\cmark} & \multirow{1}{*}{\cmark} & \multirow{1}{*}{\xmark} & 0.664 & 0.299 & 0.963 & 0.961  \\
    
  \end{tabular}
}
\caption{
    \textbf{
    Equivariance Generalizes well in One-shot Settings
    } For simplicity but aligned with~\cref{tab:benchmark}, ``Inv'' denotes Invariance Features, ``Equiv'' denotes Equivariance Features, ``XYZ'' denotes XYZ-Positions. \cref{fig:equiv} shows the directional error (left), and predicted inner body points (right).
}
\label{tab:equiv}
\end{table}

\pheading{Sparse Marker vs. Dense Correspondence.}
In~\cref{tab:ablation}, the comparison between Ours and Ours-C demonstrates the strength of our sparse marker design. To clarify, we first use tightness vectors to find the mapping between the scan mesh and the SMPL surface. Then, each position on the posed SMPL surface corresponds to a position in canonical SMPL space, which serves as the \textbf{Dense Correspondence}. For Ours, by mapping dense points to sparse markers and aggregating them with confidence, we create a voting strategy that enhances robustness against low-confidence outliers. This approach is more effective than dense prediction, which can struggle with outliers and local minima. Our method achieves significant improvements: \textcolor{ForestGreen}{$13.7\%$} lower V2V error and \textcolor{ForestGreen}{$36.5\%$} reduced MPJPE on CAPE, along with an error decrease of \textcolor{ForestGreen}{$15.1\%$} and \textcolor{ForestGreen}{$23.9\%$} on \ddress. These consistent gains validate our use of sparse markers over dense correspondences. \Cref{fig:marker_corr} shows their differences; incorrect dense correspondences can misguide optimization, leading to skewed body part rotations (\eg, hands, forearms, head), while our sparse marker strategy remains robust through its weighted voting mechanism.

\pheading{Tightness Vector vs. Tightness Scalar.}
In~\cref{tab:ablation}, we compare the tightness vector (Ours-C) and scalar (Ours-D) under a dense correspondence setting. The model performs better without the tightness vector on the CAPE dataset, while the opposite is true for \ddress. For tight clothing with minimal cloth-to-body tightness, the scalar-only (Ours-D) works well for SMPL optimization. However, incorrect direction estimates -- like being flipped -- can disrupt fitting due to sensitivity to outliers. For loose garments, the tightness scalar lacks cues for inner points, making accurate direction crucial. 
Thus, when fit styles are uncertain, especially with loose clothing, introducing the tightness vector is advisable.

\pheading{w/ Equiv vs. w/o Equiv.}
In \cref{tab:ablation,tab:equiv}, we assess how equivariant features affect direction prediction at two scales:

\textbf{1) Full Training Set:} Compare our method with Ours-A (xyz positions) and Ours-B (xyz+invariance) as inputs to the \emph{Point Transformer}. All three methods perform similarly on CAPE, but our approach has a slight edge. The gap widens on the 4D-Dress dataset, showing improvements of \textcolor{ForestGreen}{$4.6\% \sim 16.0\%$} in V2V and \textcolor{ForestGreen}{$1.6\% \sim 15.1\%$} in MPJPE. This suggests that equivariance is the key to addressing hard cases, such as loose garments and significant pose variations. According to hypothesis proposed in~\cref{subsec:vector}, invariance feature inherently conflicts with Direction prediction. The poor performance of Ours-E further demonstrates this point.

\textbf{2) One-shot Settings:} To further illustrate the out-of-distribution (OOD) generalization of equivariant features, we randomly sampled one frame from each sequence to create minimal training sets ($0.66\%$ of the 4D-Dress train split and $2.1\%$ of the CAPE train split). We train Ours, Ours-A, and Ours-B on them and evaluate on the same validation sets as in \cref{tab:ablation}. In \cref{fig:equiv}, the visualization shows scan points colored by angular error (left) and predicted inner points (right). Our method shoots inner points in the correct directions in most areas and the inner points closely approximate the inner body shape, while Ours-A and Ours-B fail completely. \cref{tab:equiv} presents mean and median angular errors (\cref{para:metrics}) between predicted and ground truth directions, highlighting our method's advantages, with notable median angular error improvements of \textcolor{ForestGreen}{$82.1\% \sim 89.8\%$} on CAPE and \textcolor{ForestGreen}{$67.2\% \sim 68.1\%$} on 4D-Dress. These results demonstrate that equivariant features enable robust direction prediction and strong OOD generalization with limited data.

\begin{table}[t]
\scriptsize
\centering{
  \begin{tabular}{c|cc|cc}
    Methods & \multicolumn{2}{c|}{CAPE~\cite{ma2020cape}} & \multicolumn{2}{c}{\ddress~\cite{wang20244ddress}}\\
     &  V2V $\downarrow$ & MPJPE $\downarrow$ & V2V $\downarrow$ & MPJPE $\downarrow$\\
     \shline
     \multicolumn{5}{c}{Tightness-agnostic} \\
     \hline
     NICP~\cite{marin24nicp} & 1.726 & 1.343 & 4.754 & 3.654 \\
     NICP$\dagger$~\cite{marin24nicp} & 1.245  & 1.051  & 4.738  & 3.729 \\
     \hline
     \multicolumn{5}{c}{Tightness-aware} \\
     \hline
     Ours & 1.647 & 0.922 & \textbf{1.939} & \textbf{1.116} \\
     Ours$\dagger$ & \textbf{1.017} & \textbf{0.883}  & 3.474  & 2.849  \\
  \end{tabular}
}
\caption{
    \textbf{Chamfer-based Post-refinement.}  We adopt the best tightness-agnostic approach, NICP~\cite{marin24nicp}, and our \modelname, to further analyze the effectiveness of chamfer-based post-refinement. Notably, $\dagger$ denotes the method w/ chamfer-based post-refinement. The results show that post-refinement \textit{improves} performance on tight clothing (CAPE~\cite{ma2020cape}) but \textit{degrades} it for loose clothing (\ddress~\cite{wang20244ddress}). Therefore, from application perspective, when clothing styles or fit are uncertain, including the ``tightness-vector'' and excluding the ``post-refinement'' will yield plausible results.}

\label{tab:chamfer}
\end{table}

\pheading{Chamfer-based Post-refinement.} As we have discussed in~\cref{sec:related} and illustrated in \cref{fig:reg_fit}, for minimal clothing scenarios like CAPE~\cite{ma2020cape}, aligning with the \textit{\textcolor{blue}{outer surface}} enhances fitting, but it may harm accuracy for loose clothing, such as \ddress~\cite{wang20244ddress}, where clothing significantly deviates from the \textit{\textcolor{red}{underlying body}}.
The results in~\cref{tab:chamfer} echo our assumption. We ablate the necessity of ``Chamfer-based Post-refinement'' on NICP~\cite{marin24nicp} and our method, and find that both benefit from such a refinement step on the CAPE dataset, yet our post-refined results still outperform NICP due to better pose initialization. In contrast, for loose clothing \ddress, the significant displacement between body and cloth will dominate the post-refinement, inflating the fitted body (see NICP's results in~\cref{fig:reg_fit}), ultimately worsening fitting results, as shown in ~\cref{tab:chamfer} (\ddress, Ours vs. Ours$\dagger$).

\pheading{Shape Accuracy.}\label{para:shape} The body fitting errors arise from pose and shape errors. The joint error MPJPE in~\cref{tab:benchmark} confirms pose accuracy. We evaluate ``shape accuracy'' in optimization-related methods (\ie, \ipnet, PTF, and NICP). \arteq~\cite{feng2023generalizing} is excluded for fair comparison as it has access to the \gt shape parameters during training. In \cref{fig:consistency}, we calculate the Mean Absolute Error (MAE) across \ddress dataset, for the first three principal shape parameters, $\shapecoeff_{[:3]} \in \real^{3}$, which feature key shape information. Our method achieves an average improvement of \textcolor{ForestGreen}{$49.9\%$}. This performance edge over tightness-aware methods like \ipnet and PTF indicates that using tightness vectors to identify the inner body surface, combined with sparse markers, yields more accurate inner body predictions than triple-layer occupancy regression~\cite{bhatnagar2020ipnet}, which is also supported in~\cref{tab:benchmark}, even without any SMPL fitting step, Chamfer distance (CD) of \modelname is still smaller than PTF and \ipnet.

\begin{figure*}[th]
    \centering{
    \includegraphics[trim=000mm 000mm 000mm 000mm, clip=true, width=\linewidth]{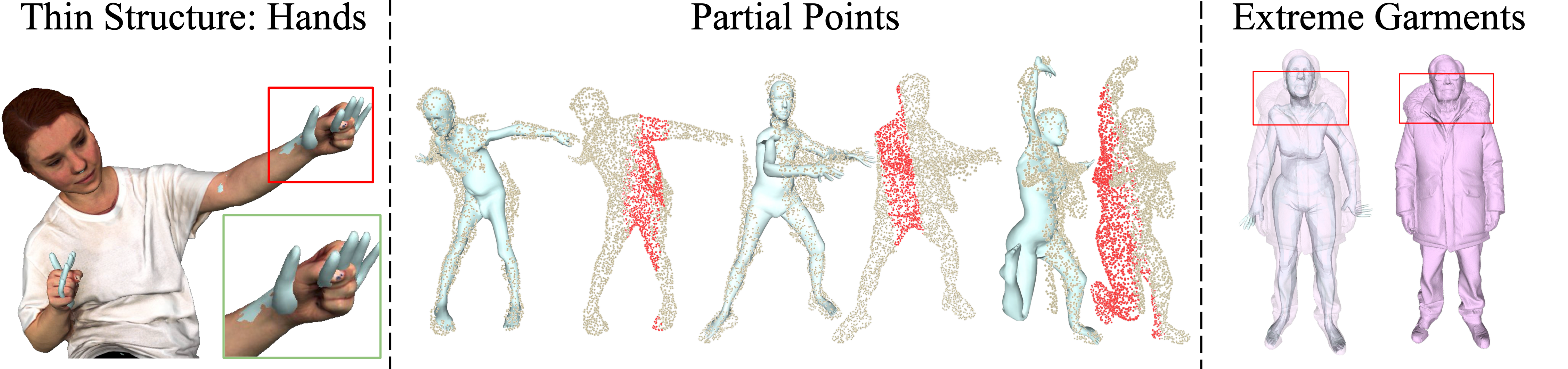}
    \caption{\textbf{Failure Cases.} ``Thin Structure'' shows that ETCH cannot perfectly fit thin structures such as hands into the scan, since sparse markers do not cover these areas; In ``Partial Points'', \textcolor{red}{Red} points indicate missing partial regions. It leads to incomplete information, which in turn results in failed SMPL fitting; ``Extreme Garments'' demonstrates a scan featuring a thick fur-trimmed hood, which is rare in the training data but present in the synthetic data. Although ETCH generalizes to different garments, it still struggles with scans that have extremely abnormal appearances or clothing.
    }
    \label{fig:limitation}}
\end{figure*}

\section{Limitations and Future Works}
\label{sec:limitation}

While \modelname shows strong performance compared to existing works, it has limitations: \textbf{1) Partial Inputs} -- As the middle part of ~\cref{fig:limitation} shows, the reliance on sparse markers means that missing point clouds prevent marker capture and lead to fitting failures. \textbf{2) Thin Structures} -- The current marker setup does not cover facial landmarks or fingers, and extending it to \smplx~\cite{SMPL:2015} is non-trivial, as the reception field needs to adapt for full-body, face and hands (refer to the left panel of~\cref{fig:limitation}) within a unified framework, potentially requiring relaxed tightness modeling when skin is detected. \textbf{3) Scalability} -- Although we excel in one-shot settings~\cref{tab:equiv} and achieve \sota results on CAPE and \ddress, performance gaps narrow with larger data volumes. It remains unclear if performance will plateau at billion-level scan-body pairs, as high-fidelity clothed datasets are scarce and costly. Synthetic~\cite{Black_CVPR_2023,zou2023cloth4d,yang2023synbody,GarmentCode2023,GarmentCodeData:2024,li2024garmentdreamer} or body-guided generated humans~\cite{liao2023tada,xiu2024puzzleavatar,li2024pshuman,zhuang2024idolinstant} could be alternatives, but domain gap issues remain. 

We will leave all above for future work. In addition, it is straightforward to extend the ``tightness vector'' to the 2D domain; related topics have already been explored in~\cite{dai2023cloth2body,guan20102d}, and Image2Body\footnote{\href{https://huggingface.co/spaces/yeq6x/Image2Body_gradio}{https://huggingface.co/spaces/yeq6x/Image2Body\_gradio}}.

\section{Conclusion}
\label{sec:conclusion}

\modelname generalizes the fitting of the \textcolor{red}{\textit{underlying body}} to diverse \textcolor{blue}{\textit{clothed humans}} through a novel framework powered by Equivariant Tightness Vector and Marker-Aggregation Optimization. We are neither the first to disentangle the clothing layer from clothed scans~\cite{bhatnagar2020ipnet,wang2021ptf} nor the pioneers in modeling tightness~\cite{chen2021tightcap,guan20102d}, but we are the first to do so using equivariant vectors, significantly outperforming previous works across various datasets and metrics. While there is still room for improvement as detailed in~\cref{sec:limitation}, 
\modelname paves the way for truly generalizable cloth-to-body fitting that is robust to any body pose and shape, garment type, and non-rigid dynamics, achieving fundamental improvements. 
We believe \modelname will reshape new perspectives on not only body fitting, but also garment refitting, real-to-sim, and biomechanics analysis. However, this technique could be misused by the porn industry and pose a threat to human privacy; therefore, we will release the packages solely for non-commercial scientific research purposes under a strict license to regulate user behavior.

\bigskip
\noindent\rule[3.0ex]{\linewidth}{1.5pt}

\noindent\textbf{Acknowledgments.} 
We thank \textit{Marilyn Keller} for the help in Blender rendering, \textit{Brent Yi} for fruitful discussions, \textit{Ailing Zeng} and \textit{Yiyu Zhuang} for HuGe100K dataset, \textit{Jingyi Wu} and \textit{Xiaoben Li} for their help during rebuttal, and the members of \textit{Endless AI Lab} for their help and discussions. This work is funded by the Research Center for Industries of the Future (RCIF) at Westlake University, the Westlake Education Foundation.
\textit{Yuliang Xiu} also received funding from the Max Planck Institute for Intelligent Systems.

\medskip

\noindent\textbf{Disclosure}
While MJB is a co-founder and Chief Scientist at Meshcapade, his research in this project was performed solely at, and funded solely by, the Max Planck Society.

\clearpage

\clearpage
\clearpage

{
    \small
    \bibliographystyle{config/ieeenat_fullname}
    \bibliography{main}
}

\end{document}